\documentclass[runningheads]{llncs}
\usepackage{graphicx}
\usepackage{amsmath,amssymb} 
\usepackage{color}
\usepackage[width=122mm,left=12mm,paperwidth=146mm,height=193mm,top=12mm,paperheight=217mm]{geometry}
\usepackage[font={small}]{caption} %
\usepackage[font={small}]{subcaption}%
\usepackage[dvipsnames]{xcolor}
\usepackage{tabularx}
\usepackage{multirow}
\usepackage{booktabs}
\usepackage{wrapfig}
\usepackage{comment}
\usepackage{enumitem}
\usepackage{mathptmx}
\usepackage[11pt]{moresize}
\usepackage[rightcaption]{sidecap}
\usepackage{anyfontsize}
\usepackage{hyperref}
\hypersetup{
    colorlinks=true,
    linkcolor=red,
    filecolor=red,      
    urlcolor=red,
}
\usepackage{cite}

\usepackage{tikz}
\usetikzlibrary{decorations.pathreplacing}
\usetikzlibrary{shapes.callouts}
\usetikzlibrary{arrows}
\usetikzlibrary{chains, fit, quotes, automata, arrows}
\usetikzlibrary{matrix,positioning}
\usepackage{mathtools}

\def\etal{\emph{et al. }}

\def\Fig{Fig. }

\begin{document}
\pagestyle{headings}
\mainmatter
\def\ECCV18SubNumber{2342}  

\title{ESPNet: Efficient Spatial Pyramid of Dilated Convolutions for Semantic Segmentation} 

\titlerunning{ESPNet: Efficient Spatial Pyramid of Dilated Convolutions for Semantic Segmentation}

\authorrunning{Mehta \etal}

\author{Sachin Mehta\inst{1} \and Mohammad Rastegari\inst{2} \and Anat Caspi\inst{1} \and \\ Linda Shapiro\inst{1}  \and  Hannaneh Hajishirzi\inst{1}}
\institute{University of Washington, Seattle, WA, USA \\
	\email{\{sacmehta, caspian, shapiro, hannaneh\}@cs.washington.edu} \\
	\and Allen Institute for AI and XNOR.AI, Seattle, WA, USA \\
	\email{mohammadr@allenai.org}\\
\textbf{\large Source code:} \large{\href{https://github.com/sacmehta/ESPNet}{https://github.com/sacmehta/ESPNet}}
}

\maketitle

\begin{abstract}
We introduce a fast and efficient convolutional neural network, ESPNet, for semantic segmentation of high resolution images under resource constraints. ESPNet is based on a new convolutional module, efficient spatial pyramid (ESP), which is efficient in terms of computation, memory, and power. ESPNet is 22 times faster (on a standard GPU) and 180 times smaller than the state-of-the-art semantic segmentation network PSPNet \cite{zhao2017pyramid}, while its category-wise accuracy is only 8\% less. We evaluated ESPNet on a variety of semantic segmentation datasets including  Cityscapes,  PASCAL VOC, and a breast biopsy whole slide image dataset. Under the same constraints on memory and computation, ESPNet outperforms all the current efficient CNN networks such as MobileNet \cite{howard2017mobilenets}, ShuffleNet \cite{zhang2017shufflenet}, and ENet \cite{paszke2016enet} on both standard metrics and our newly introduced performance metrics that measure efficiency on edge devices. Our network can process high resolution images at a rate of 112 and 9 frames per second on a standard GPU and edge device, respectively.   
\end{abstract}

\section{Introduction}
Deep convolutional neural network (CNN) models have achieved high accuracy in visual scene understanding tasks \cite{zhao2017pyramid,he2014spatial,chen2016deeplab}. While the accuracy of these networks has improved with their increase in depth and width, large networks are slow and power hungry. This is especially problematic on the computationally heavy task of semantic segmentation \cite{ess2009segmentation,geiger2013vision,cordts2016cityscapes,menze2015object,franke2013making,xiang2017rnn,kundu2014joint}.
For example, PSPNet \cite{zhao2017pyramid} has 65.7 million parameters and runs at about 1 FPS while discharging the battery of a standard laptop at a rate of 77 Watts. Many advanced real-world applications, such as self-driving cars, robots, and augmented reality, are sensitive and demand on-line processing of data locally on edge devices. These accurate networks require enormous resources and are not suitable for edge devices, which have limited energy overhead, restrictive memory constraints, and reduced computational capabilities.

\begin{figure}[t!]
\centering
\begin{subfigure}[b]{0.48\columnwidth}
\centering
\includegraphics[width=\columnwidth]{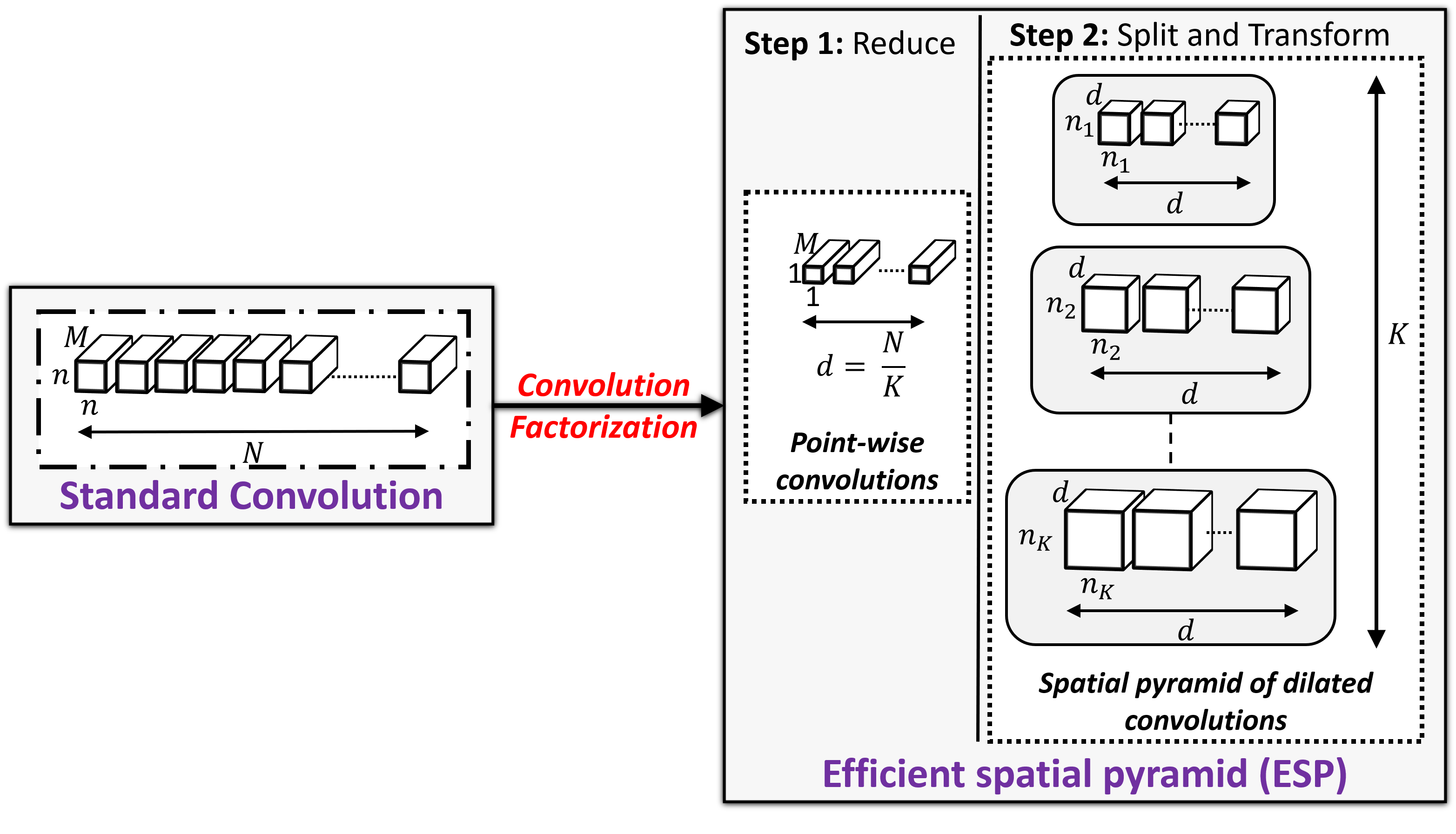}
\caption{}
\label{fig:kernelVis}
\end{subfigure}
\hfill
\begin{subfigure}[b]{0.48\columnwidth}
\centering
\resizebox{\columnwidth}{!}{
\hspace{-90pt}\input{Digrams.tikz}\esppDegRes
}
\caption{}
\label{fig:sppBlocks}
\end{subfigure}
\caption{(a) The standard convolution layer is decomposed into point-wise convolution and spatial pyramid of dilated convolutions to build an efficient spatial pyramid (ESP) module. (b) Block diagram of ESP module. The large effective receptive field of the ESP module introduces gridding artifacts, which are removed using hierarchical feature fusion (HFF). A skip-connection between input and output is added to improve the information flow. See Section \ref{sec:espnetArc} for more details. Dilated convolutional layers are denoted as (\# input channels, effective kernel size, \# output channels). The effective spatial dimensions of a dilated convolutional kernel are $n_k \times n_k$, where $n_k = (n-1)2^{k-1} + 1,\  k=1, \cdots, K$. Note that only $n\times n$ pixels participate in the dilated convolutional kernel. In our experiments $n=3$ and $d=\frac{M}{K}$.}
\label{fig:espDesc}
\end{figure}

Convolution factorization has demonstrated its success in reducing the computational complexity of deep CNNs (e.g. Inception\cite{szegedy2015going,szegedy2016rethinking,SzegedyIV16InceptionV4},  ResNext \cite{xie2017aggregated}, and Xception \cite{chollet2016xception}). We introduce an efficient convolutional module, ESP (efficient spatial pyramid),  which is based on the convolutional factorization principle (\Fig \ref{fig:espDesc}). Based on these ESP modules, we introduce an efficient network structure, ESPNet, that can be easily deployed on resource-constrained edge devices. ESPNet is \textit{fast}, \textit{small}, \textit{low power}, and \textit{low latency}, yet still preserves segmentation accuracy.

ESP is based on a convolution factorization principle that decomposes a standard convolution into two steps: (1) \textbf{\textit{point-wise convolutions}} and (2) \textbf{\textit{spatial pyramid of dilated convolutions}}, as shown in \Fig \ref{fig:espDesc}. The point-wise convolutions help in reducing the computation, while the spatial pyramid of dilated convolutions re-samples the feature maps to learn the representations from large effective receptive field. We show that our ESP module is more efficient than other factorized forms of convolutions, such as Inception \cite{szegedy2015going,szegedy2016rethinking,SzegedyIV16InceptionV4} and  ResNext \cite{xie2017aggregated}. Under the same constraints on memory and computation, ESPNet outperforms  MobileNet \cite{howard2017mobilenets} and ShuffleNet \cite{zhang2017shufflenet} (two other efficient networks that are built upon the factorization principle). We note that existing spatial pyramid methods (e.g. the atrous spatial pyramid module in \cite{chen2016deeplab}) are computationally expensive and cannot be used at different spatial levels for learning the representations. In contrast to these methods, ESP is computationally efficient and can be used at different spatial levels of a CNN network. Existing models based on dilated convolutions \cite{zhao2017pyramid,chen2016deeplab,yu2015multi,yu2017dilated} are large and inefficient, but our ESP module generalizes the use of dilated convolutions in a novel and efficient way.   

To analyze the performance of a CNN network on edge devices, we introduce several new performance metrics, such as sensitivity to GPU frequency and warp execution efficiency.
To showcase the power of ESPNet, we evaluate our model on one of the most expensive tasks in AI and computer vision: semantic segmentation. ESPNet is empirically demonstrated to be more accurate, efficient, and fast than ENet \cite{paszke2016enet}, one of the most power-efficient semantic segmentation networks, while learning a similar number of parameters. Our results also show that ESPNet learns generalizable representations and outperforms ENet \cite{paszke2016enet} and another efficient network ERFNet \cite{romera2018erfnet} on the unseen dataset. ESPNet can process a \textit{high resolution RGB image} at a rate of 112 frames per second (FPS) on a high-end GPU, 21 FPS on a laptop, and 9 FPS on an edge device\footnote{We used a desktop with NVIDIA TitanX GPU, a laptop with GTX-960M GPU, and NVIDIA Jetson TX2 as an edge device. See \textbf{Appendix \ref{sec:hardwareDetailsSup}} for more details.}.  

\section{Related Work}
Multiple different techniques, such as convolution factorization, network compression, and low-bit networks, have been proposed to speed up convolutional neural networks. We, first, briefly describe these approaches and then provide a brief overview of CNN-based semantic segmentation.

\noindent \textbf{Convolution factorization:} Convolutional factorization decomposes the convolutional operation into multiple steps to reduce the computational complexity. This factorization has successfully shown its potential in reducing the computational complexity of deep CNN networks (e.g. Inception \cite{szegedy2015going,szegedy2016rethinking,SzegedyIV16InceptionV4}, factorized network \cite{jin2014flattened}, ResNext \cite{xie2017aggregated}, Xception \cite{chollet2016xception}, and MobileNets \cite{howard2017mobilenets}). ESP modules are also built on this factorization principle. The ESP module decomposes a convolutional layer into a point-wise convolution and spatial pyramid of dilated convolutions. This factorization helps in reducing the computational complexity, while simultaneously allowing the network to learn the representations from a large effective receptive field.

\noindent \textbf{Network Compression:} Another approach for building efficient networks is compression. These methods use techniques such as hashing \cite{chen2015compressing}, pruning \cite{han2015deep}, vector quantization \cite{wu2016quantized}, and shrinking \cite{zhao2017icnet,jaderberg2014speeding} to reduce the size of the pre-trained network.

\noindent \textbf{Low-bit networks:} Another approach towards efficient networks is low-bit networks, which quantize the weights to reduce the network size and complexity (e.g. \cite{rastegari2016xnor,Hwang2014fixed,courbariaux2016binarized,hubara2016quantized}). 

\noindent \textbf{Sparse CNN:} To remove the redundancy in CNNs, sparse CNN methods, such as sparse decomposition \cite{liu2015sparse}, structural sparsity learning \cite{wen2016learning}, and dictionary-based method \cite{bagherinezhad2017lcnn}, have been proposed. 

We note that compression-based methods, low-bit networks, and sparse CNN methods are equally applicable to ESPNets and are complementary to our work.

\noindent \textbf{Dilated convolution:} Dilated convolutions \cite{holschneider1990real} are a special form of standard convolutions in which the effective receptive field of kernels is increased by inserting zeros (or holes) between each pixel in the convolutional kernel. For a $n \times n$ dilated convolutional kernel with a dilation rate of $r$, the effective size of the kernel is $\left[ (n-1) r + 1\right]^2$. The dilation rate specifies the number of zeros (or holes) between  pixels. However, due to dilation, only $n \times n$ pixels participate in the convolutional operation, reducing the computational cost while increasing the effective kernel size.

Yu and Koltun \cite{yu2015multi} stacked dilated convolution layers with increasing dilation rate to learn contextual representations from a large effective receptive field. A similar strategy was adopted in \cite{yu2017dilated,mehta2017learning,wang2017understanding}. Chen \etal \cite{chen2016deeplab} introduced an atrous spatial pyramid (ASP) module. This module can be viewed as a parallelized version of \cite{chen2016deeplab}. These modules are computationally inefficient (e.g. ASPs have high memory requirements and learn many more parameters; see Section \ref{ssec:relationship}). Our ESP module also learns multi-scale representations using dilated convolutions in parallel; however, it is computationally efficient and can be used at any spatial level of a CNN network. 

\noindent \textbf{CNN for semantic segmentation:} Different CNN-based segmentation networks have been proposed, such as multi-dimensional recurrent neural networks \cite{mrnn:eke}, encoder-decoders \cite{paszke2016enet,romera2018erfnet,badrinarayanan2017segnet,ronneberger2015u}, hypercolumns \cite{hariharan2015hypercolumns}, region-based representations \cite{dai2015convolutional:eke,Caesar2016:eke}, and cascaded networks \cite{lin2017refinenet}. Several supporting techniques along with these networks have been used for achieving high accuracy, including ensembling features \cite{chen2016deeplab}, multi-stage training \cite{long2015fully}, additional training data from other datasets \cite{zhao2017pyramid,chen2016deeplab}, object proposals \cite{Noh2015:eke}, CRF-based post processing \cite{chen2016deeplab}, and pyramid-based feature re-sampling \cite{zhao2017pyramid,he2014spatial,chen2016deeplab}.

\noindent \textbf{\textit{Encoder-decoder networks:}} Our work is related to this line of work. The encoder-decoder networks first learn the representations by performing convolutional and down-sampling operations. These representations are then decoded by performing up-sampling and convolutional operations. ESPNet first learns the encoder and then attaches a \textit{light-weight decoder} to produce the segmentation mask. This is in contrast to existing networks where the decoder is either an exact replica of the encoder (e.g. \cite{badrinarayanan2017segnet}) or  is relatively small (but not light weight) in comparison to the encoder (e.g. \cite{paszke2016enet,romera2018erfnet}). 

\noindent \textbf{\textit{Feature re-sampling methods:}} The feature re-sampling methods re-sample the convolutional feature maps at the same scale using different pooling rates \cite{zhao2017pyramid,he2014spatial} and kernel sizes \cite{chen2016deeplab} for efficient classification. Feature re-sampling is computationally expensive and is performed just before  the classification layer to learn scale-invariant representations. We introduce a computationally efficient convolutional module that allows feature re-sampling at different spatial levels of a CNN network.

\section{ESPNet}
\label{sec:espnetArc}
This section  elaborates on the details of ESPNET and describes the core ESP module on which it is built. We compare ESP modules with similar CNN modules, such as Inception \cite{szegedy2015going,szegedy2016rethinking,SzegedyIV16InceptionV4}, ResNext \cite{xie2017aggregated}, MobileNet\cite{howard2017mobilenets}, and ShuffleNet\cite{zhang2017shufflenet} modules. 

\subsection{ESP module}
ESPNet is based on efficient spatial pyramid (ESP) modules, which are a factorized form of convolutions that decompose a standard convolution into a point-wise convolution and a spatial pyramid of dilated convolutions (see \Fig \ref{fig:kernelVis}). The point-wise convolution in the ESP module applies a $1\times 1$ convolution to project high-dimensional feature maps onto a low-dimensional space. The spatial pyramid of dilated convolutions then re-samples these low-dimensional feature maps using $K$, $n\times n$ dilated convolutional kernels simultaneously, each with a dilation rate of $2^{k-1}$, $k = \{1, \cdots, K\}$. This factorization drastically reduces the number of parameters and the memory required by the ESP module, while preserving a large effective receptive field $\left[ (n-1) 2^{K-1} + 1\right]^2$. This pyramidal convolutional operation is called a spatial pyramid of dilated convolutions, because each dilated convolutional kernel learns weights with different receptive fields and so resembles a spatial pyramid.  

A standard convolutional layer takes an input feature map $\mathbf{F}_i \in \mathbb{R}^{W \times H \times M}$ and applies $N$ kernels $\mathbf{K} \in \mathbb{R}^{m \times n\times M}$ to produce an output feature map $\mathbf{F}_o \in \mathbb{R}^{W \times H \times N}$, where $W$ and $H$ represent the width and height of the feature map, $m$ and $n$ represent the width and height of the kernel, and $M$ and $N$ represent the number of input and output feature channels. For the sake of simplicity, we will assume that $m=n$. A standard convolutional kernel thus learns $n^2MN$ parameters. These parameters are \textit{multiplicatively} dependent on the spatial dimensions of the $n \times n$ kernel and the number of input $M$ and output $N$ channels. 

\noindent \textbf{Width divider $K$:} To reduce the computational cost, we introduce a simple hyper-parameter $K$. The role of $K$ is to shrink the dimensionality of the feature maps uniformly across each ESP module in the network. \textit{Reduce:} For a given $K$, the ESP module first reduces the feature maps from $M$-dimensional space to $\frac{N}{K}$-dimensional space using a point-wise convolution (Step 1 in \Fig \ref{fig:kernelVis}). \textit{Split:} The low-dimensional feature maps are then split across $K$ parallel branches. \textit{Transform:} Each branch then processes these feature maps simultaneously using $n\times n$ dilated convolutional kernels with different dilation rates given by $2^{k-1},\  k=\{1, \cdots, K-1\}$ (Step 2 in \Fig \ref{fig:kernelVis}). \textit{Merge:} The output of these $K$ parallel dilated convolutional kernels is then concatenated to produce an $N$-dimensional output feature map\footnote{In general, $\frac{N}{K}$ may not be a perfect divisor, and therefore concatenating $K$, $\frac{N}{K}$-dimensional feature maps would not result in an $N$-dimensional output. To handle this, we use $\left(N - (K - 1) \lfloor\frac{N}{K}\rfloor\right)$ kernels with a dilation rate of $2^0$ and $\lfloor\frac{N}{K}\rfloor$ kernels for each dilation rate $2^{k-1}$ for $k=\{2, \cdots, K\}$.}. \Fig \ref{fig:sppBlocks} visualizes the \textit{reduce-split-transform-merge} strategy used in ESP modules.

The ESP module has $\frac{MN}{K} + \frac{(nN)^2}{K}$ parameters and its effective receptive field is $\left[ (n-1) 2^{K-1} + 1\right]^2$. Compared to the $n^2NM$ parameters of the standard convolution,  factorizing it using the two steps reduces the total number of parameters in the ESP module by a factor of $\frac{n^2 M K}{M + n^2 N}$, while increasing the effective receptive field by $\sim [2^{K-1}]^2$. For example, an ESP module learns $\sim 3.6$ times fewer parameters with an effective receptive field of $17 \times 17$ than a standard convolutional kernel with an effective receptive field of $3 \times 3$ for $n=3$, $N=M=128$, and $K=4$.

\begin{figure}[t!]
\centering
\begin{subfigure}[b]{0.22\columnwidth}
\centering
\includegraphics[height=150px]{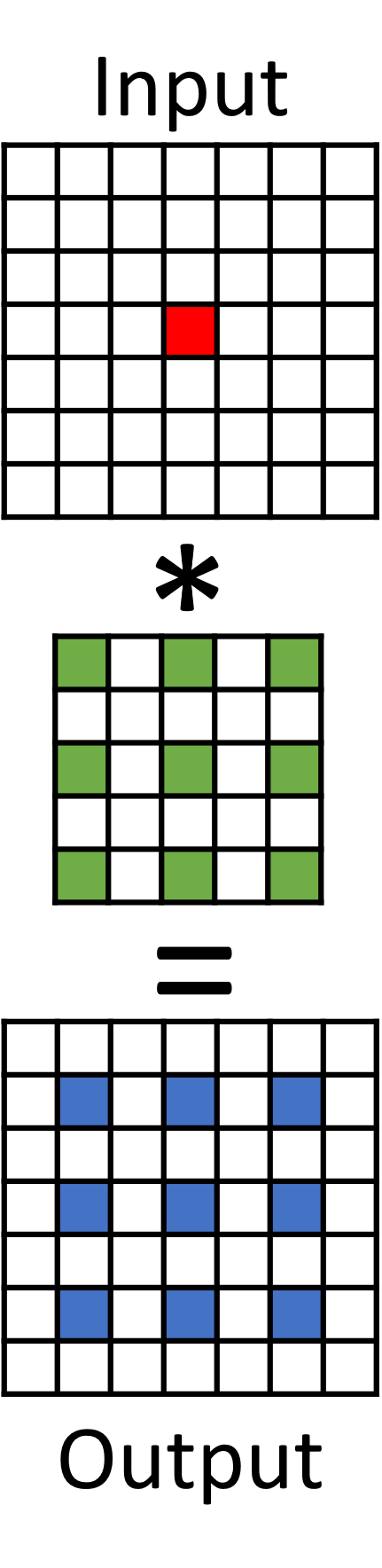}
\caption{}
\label{fig:gridArtifact}
\end{subfigure}
\hfill
\begin{subfigure}[b]{0.77\columnwidth}
  \centering
	\begin{tabular}{ccc}
    \textbf{RGB} & \textbf{without HFF} & \textbf{with HFF} \\
    \includegraphics[width=0.32\columnwidth]{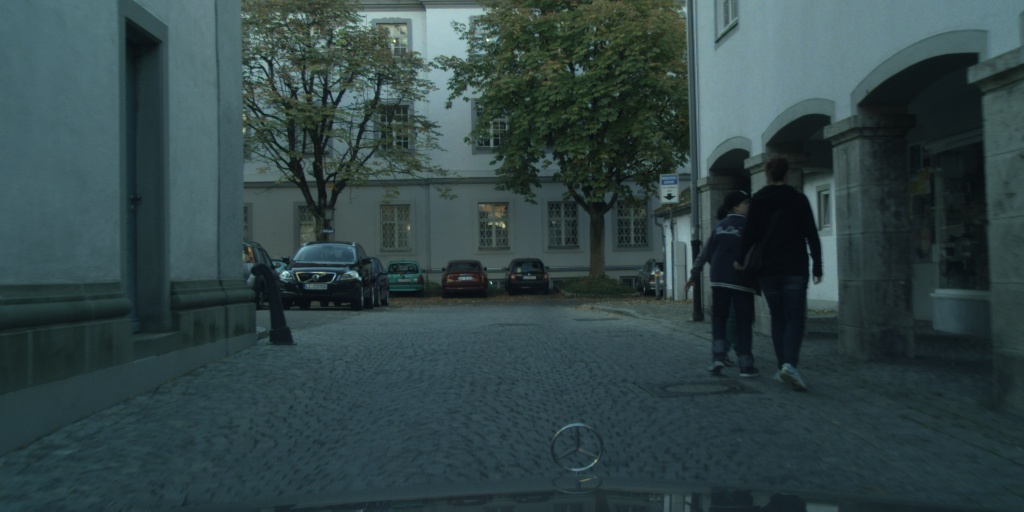} & \includegraphics[width=0.32\columnwidth]{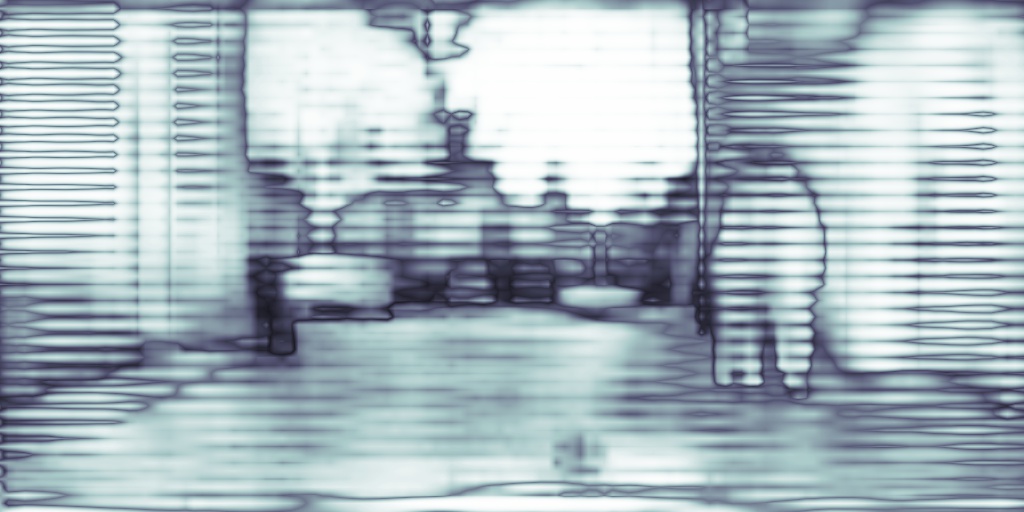} & \includegraphics[width=0.32\columnwidth]{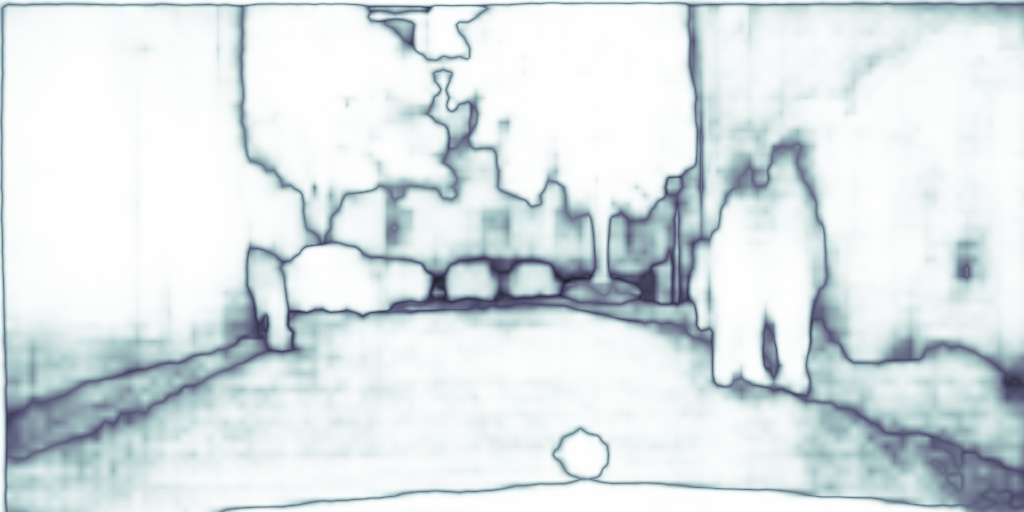}  \\
    \includegraphics[width=0.32\columnwidth]{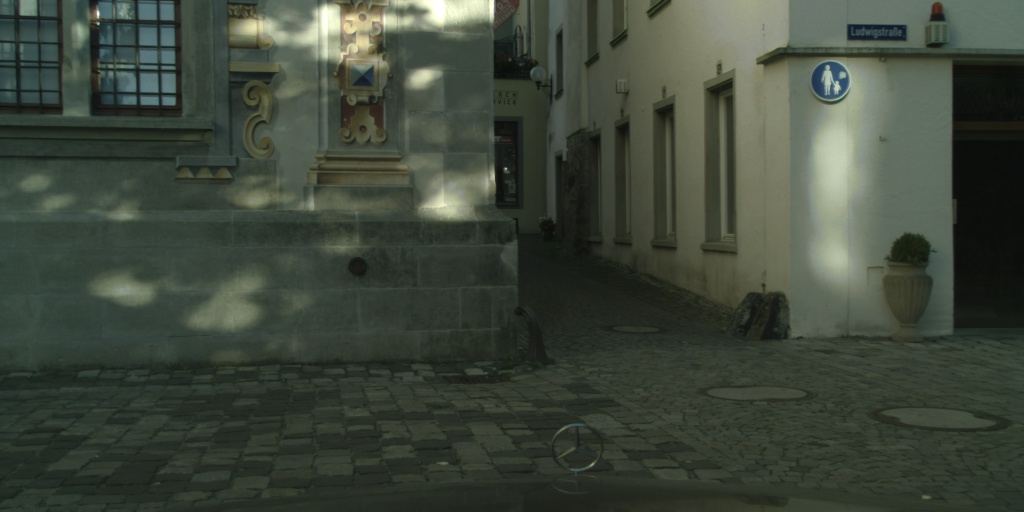} & \includegraphics[width=0.32\columnwidth]{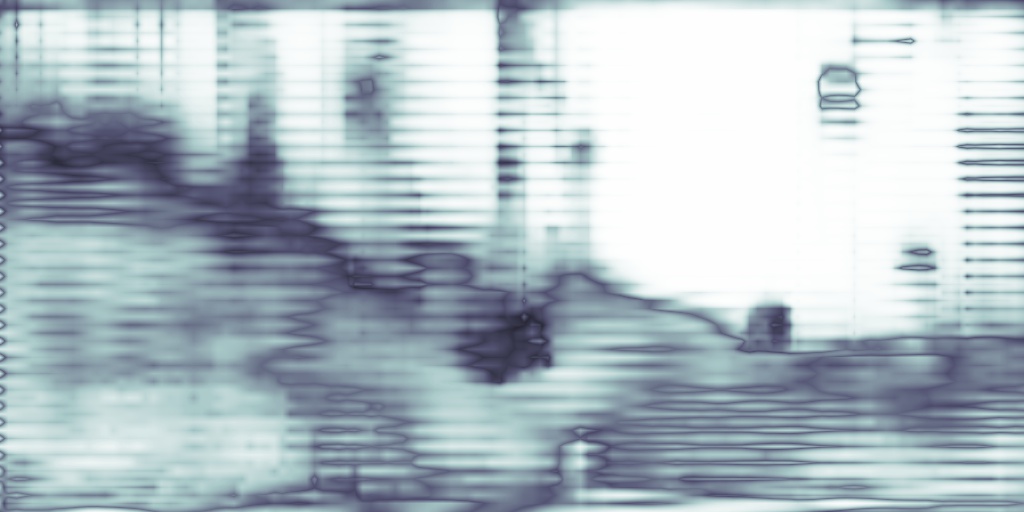} & \includegraphics[width=0.32\columnwidth]{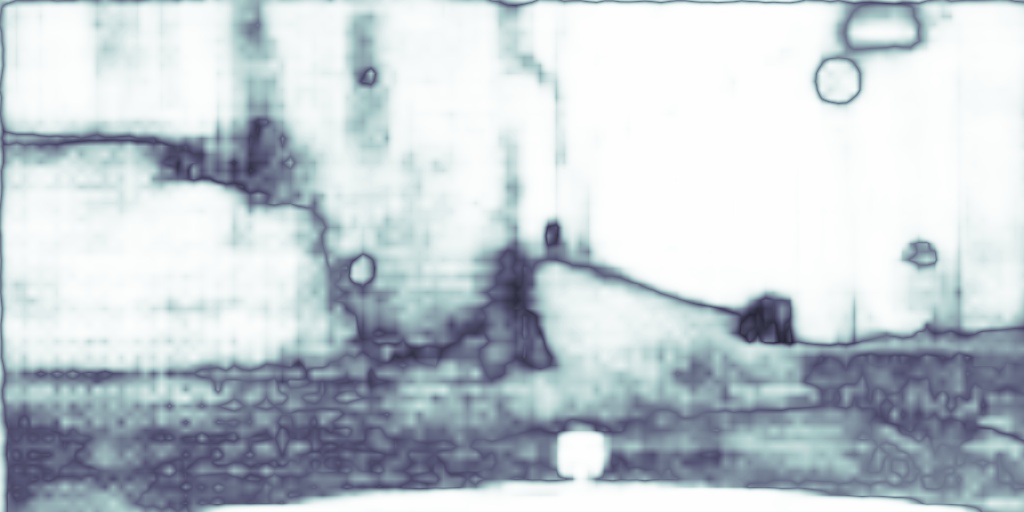}  \\
    \includegraphics[width=0.32\columnwidth]{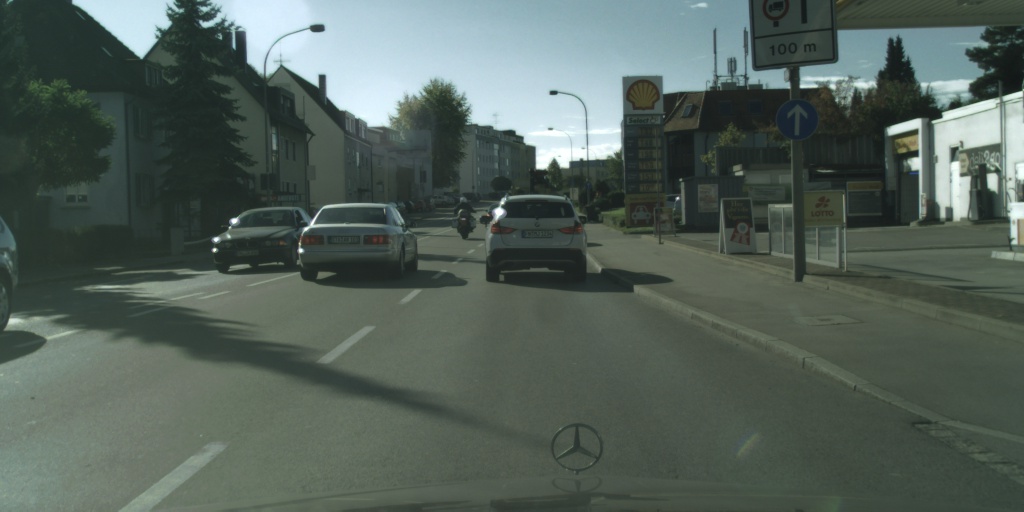} & \includegraphics[width=0.32\columnwidth]{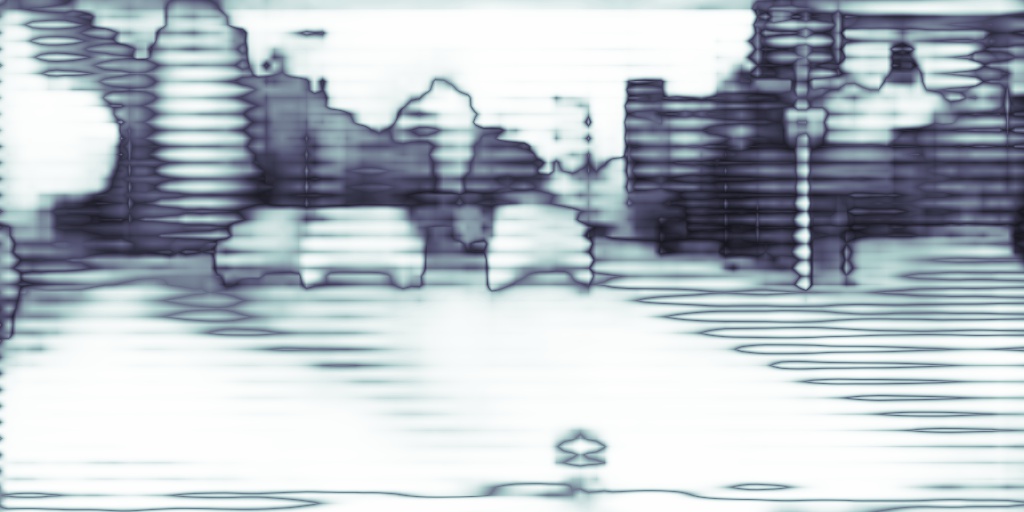} & \includegraphics[width=0.32\columnwidth]{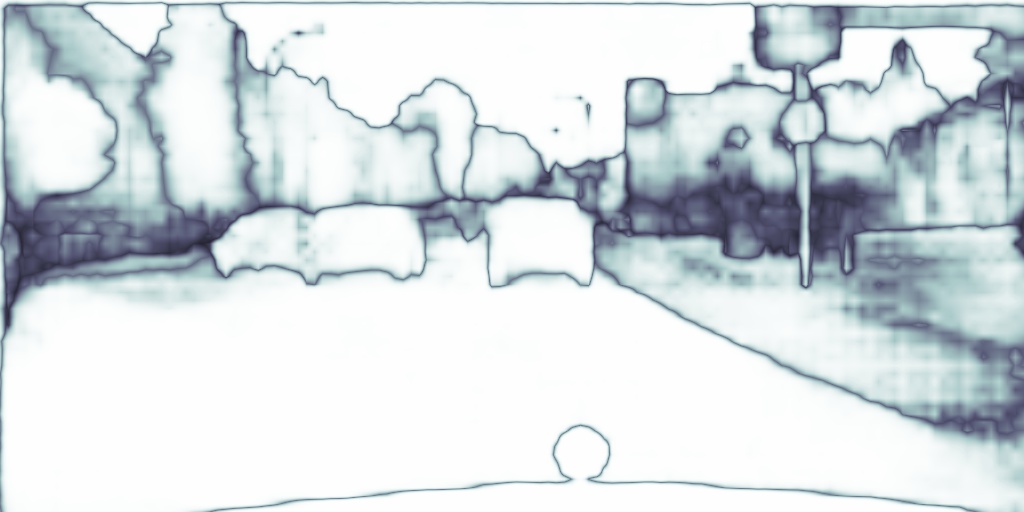}  \\
    \end{tabular}
  \caption{}
\label{fig:gridArtExm}
\end{subfigure}
\caption{(a) An example illustrating a gridding artifact with a single active pixel (red) convolved with a $3\times3$ dilated convolutional kernel with dilation rate $r=2$. (b) Visualization of feature maps of ESP modules with and without hierarchical feature fusion (HFF). HFF in ESP eliminates the gridding artifact. Best viewed in color.}
\label{fig:gridArtExmamination}
\end{figure}

\noindent \textbf{Hierarchical feature fusion (HFF) for de-gridding:} While concatenating the outputs of dilated convolutions give the ESP module a large effective receptive field, it introduces unwanted checkerboard or gridding artifacts, as shown in \Fig \ref{fig:gridArtExmamination}. To address the gridding artifact in ESP, the feature maps obtained using kernels of different dilation rates are hierarchically added before concatenating them (HFF in \Fig \ref{fig:sppBlocks}). This solution is simple and effective and does not increase the complexity of the ESP module, in contrast to  existing methods that remove the gridding artifact by learning more parameters using dilated convolutional kernels with small dilation rates \cite{yu2017dilated,wang2017understanding}. To improve the gradient flow inside the network, the input and output feature maps of the ESP module are combined using an element-wise sum \cite{he2016deep}.

\subsection{Relationship with other CNN modules}
\label{ssec:relationship}
The ESP module shares similarities with the following CNN modules.

\noindent \textbf{MobileNet module:} The MobileNet module \cite{howard2017mobilenets}, visualized in  \Fig \ref{fig:mobile}, uses a depth-wise separable convolution \cite{chollet2016xception} that factorizes a standard convolutions into depth-wise convolutions (\textit{transform}) and point-wise convolutions (\textit{expand}). It learns less parameters, has high memory requirement, and low receptive field than the ESP module. 
An extreme version of the ESP module (with $K=N$) is almost identical to the MobileNet module, differing only in the order of convolutional operations. In the MobileNet module, the spatial convolutions are followed by point-wise convolutions; however, in the ESP module, point-wise convolutions are followed by spatial convolutions. Note that the effective receptive field of an ESP module ($\left[ (n-1) 2^{K-1} + 1 \right]^2$) is higher than a MobileNet module ($\left[n \right]^2$).

\begin{figure}[b!]
\centering
\begin{subfigure}[b]{0.22\columnwidth}
    \centering
    \resizebox{\columnwidth}{!}{
    \input{Digrams.tikz}\depthWise
    }
   \caption{MobileNet}
   \label{fig:mobile}
  \end{subfigure}
  \hfill
  \begin{subfigure}[b]{0.22\columnwidth}
    \centering
    \resizebox{0.8\columnwidth}{!}{
    \input{Digrams.tikz}\shuffle
    }
   \caption{ShuffleNet}
   \label{fig:shuffle}
  \end{subfigure}
  \hfill
  \begin{subfigure}[b]{0.4\columnwidth}
    \centering
    \resizebox{\columnwidth}{!}{
    \input{Digrams.tikz}\inception
    }
   \caption{Inception}
   \label{fig:inceptionGen}
  \end{subfigure}
    \vfill
  \begin{subfigure}[b]{0.32\columnwidth}
    \centering
    \resizebox{\columnwidth}{!}{
	\input{Digrams.tikz}\resNext
    }
   \caption{ResNext}
   \label{fig:resnext}
  \end{subfigure}
  \hfill
  \begin{subfigure}[b]{0.33\columnwidth}
    \centering
    \resizebox{\columnwidth}{!}{
    \input{Digrams.tikz}\spblock
    }
   \caption{ASP}
   \label{fig:asp}
  \end{subfigure}
  \hfill
  \begin{subfigure}[b]{0.32\columnwidth}
  	\centering
  	\resizebox{\columnwidth}{!}{
  		\input{Digrams.tikz}\esppDegResA
  	}
  	\caption{ESP (same as in \Fig \ref{fig:espDesc})}
  	\label{}
  \end{subfigure}
\vfill
\begin{subfigure}[b]{\columnwidth}
\centering
\resizebox{\columnwidth}{!}{
\begin{tabular}{l|r|r|r}
\toprule
 \textbf{Module} & \textbf{\# Parameters} & \textbf{Memory (in MB)} & \textbf{Effective Receptive Field}\\
\midrule
 MobileNet & $M(n^2 + N) = 11,009$ & $(M + N)WH = 2.39$ & $\left[ n \right]^2 = 3 \times 3$\\
\hline
ShuffleNet & $\frac{d}{g}(M + N) + n^2 d = \textbf{2,180}$ & $WH(2*d + N) = 1.67$ & $\left[ n \right]^2 = 3 \times 3$\\
\hline
 Inception & $K (M d + n^2 d^2) = 28,000$ & $2K W H d = 2.39$ & $\left[ n \right]^2 = 3 \times 3$\\
\hline
 ResNext & $K (M d + d^2 n^2 + d N) = 38,000$ & $K W H (2 d + N)  = 8.37$ & $\left[ n \right]^2 = 3 \times 3$ \\
\hline
 ASP & $K M N n^2 = 4 50,000$ & $K W H N  = 5.98$ & $\left[ (n-1) 2^{K-1} + 1\right]^2 = \mathbf{33 \times 33}$\\
  \hline
 ESP (\Fig \ref{fig:sppBlocks}) & $M d + K n^2 d^2 = 20,000$ & $ W H d  (K + 1) = \mathbf{1.43} $ & $\left[ (n-1) 2^{K-1} + 1\right]^2 = \mathbf{33 \times 33}$\\
 \bottomrule
 \multicolumn{4}{l}{Here, $M=N=100$, $n=3$, $K=5$, $d=\frac{N}{K}=20$, $g=2$, and $W=H=56$.}
\end{tabular}
}
\caption{Comparison between different modules}
\label{fig:compareBlocksTheory}
\end{subfigure}
\caption{Different types of convolutional modules for comparison. We denote the layer as (\# input channels, kernel size, \# output channels). Dilation rate in (e) is indicated on top of each layer. Here, $g$ represents the number of convolutional groups in grouped convolution \cite{krizhevsky2012imagenet}. For simplicity, we only report the memory of convolutional layers in (d). For converting the required memory to bytes, we multiply it by 4 (1 float requires 4 bytes for storage).} 
\label{fig:compareBlocks}
\end{figure}

\noindent \textbf{ShuffleNet module:} The ShuffleNet module \cite{zhang2017shufflenet}, shown in \Fig \ref{fig:shuffle},  is based on the principle of \textit{reduce-transform-expand}. It is an optimized version of the bottleneck block in ResNet \cite{he2016deep}. To reduce computation, Shufflenet makes use of grouped convolutions \cite{krizhevsky2012imagenet} and depth-wise convolutions \cite{chollet2016xception}. It replaces $1\times 1$ and $3\times 3$ convolutions in the bottleneck block in ResNet with $1\times 1$ grouped convolutions and $3\times 3$ depth-wise separable convolutions, respectively.  The Shufflenet module learns many less parameters than the ESP module, but has higher memory requirements and a smaller receptive field.

\noindent \textbf{Inception module:} Inception modules \cite{szegedy2015going,szegedy2016rethinking,SzegedyIV16InceptionV4} are built on the principle of \textit{split-reduce-transform-merge}. These modules are usually heterogeneous in number of channels and kernel size (e.g. some of the modules are composed of standard and factored convolutions). In contrast to the Inception modules, ESP modules are straightforward and simple to design. For the sake of comparison, the homogeneous version of an Inception module is shown in \Fig \ref{fig:inceptionGen}. \Fig \ref{fig:compareBlocksTheory} compares the Inception module with the ESP module. ESP (1) learns fewer parameters, (2) has a low memory requirement, and (3) has a larger effective receptive field.

\noindent \textbf{ResNext module:} A ResNext module \cite{xie2017aggregated}, shown in \Fig \ref{fig:resnext}, is a parallel version of the bottleneck module in ResNet \cite{he2016deep} and is based on the principle of \textit{split-reduce-transform-expand-merge}. The ESP module is similar to ResNext in the sense that it involves branching and residual summation. However, the ESP module is more efficient in memory and parameters and has a larger effective receptive field.

\vspace{1mm}
\noindent \textbf{Atrous spatial pyramid (ASP) module:}  An ASP module \cite{chen2016deeplab}, shown in \Fig \ref{fig:asp}, is built on the principle of \textit{split-transform-merge}. The ASP module involves branching with each branch learning kernel at a different receptive field (using dilated convolutions). Though ASP modules tend to perform well in segmentation tasks due to their high effective receptive fields, ASP modules have high memory requirements and learn many more parameters. Unlike the ASP module, the ESP module is computationally efficient.

\section{Experiments}
Semantic segmentation is one of the most expensive task in AI and computer vision. To showcase the power of ESPNet, ESPNet's performance is evaluated on several datasets for semantic segmentation and compared to the state-of-the-art networks.   

\subsection{Experimental set-up}
\noindent \textbf{Network structure:} ESPNet uses ESP modules for learning convolutional kernels as well as down-sampling operations, except for the first layer which is a standard strided convolution. All layers (convolution and ESP modules) are followed by a batch normalization \cite{ioffe2015batch} and a PReLU \cite{he2015delving} non-linearity except for the last point-wise convolution, which has neither batch normalization nor non-linearity. The last layer feeds into a softmax for pixel-wise classification. 

\begin{figure}[t!]
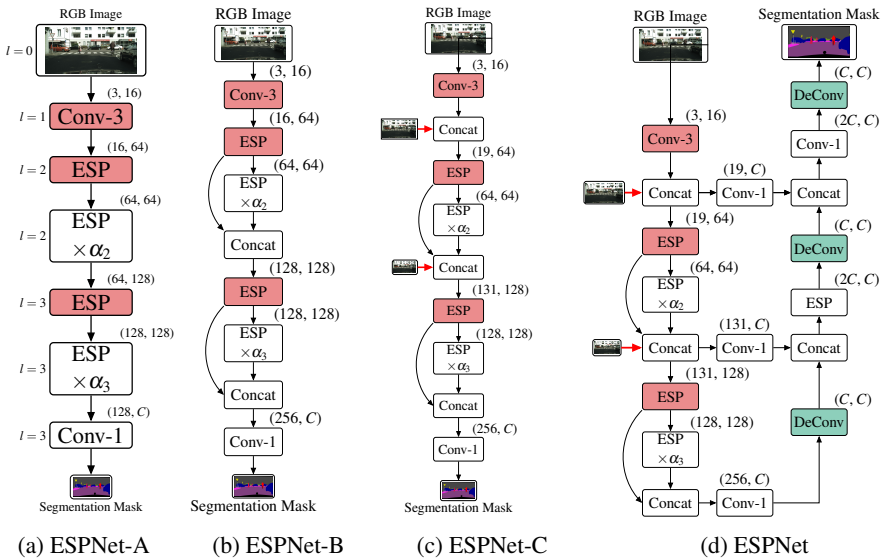

\centering
\begin{subfigure}[b]{0.18\columnwidth}
\resizebox{!}{190px}{
\hspace*{-70pt}\input{Digrams.tikz}\architectureEncA
}
\caption{ESPNet-A}
\label{fig:archA}
\end{subfigure}
\hfill
\begin{subfigure}[b]{0.2\columnwidth}
\resizebox{!}{190px}{
\hspace*{-70pt}\input{Digrams.tikz}\architectureEncB
}
\caption{ESPNet-B}
\label{fig:archB}
\end{subfigure}
\hfill
\begin{subfigure}[b]{0.2\columnwidth}
\resizebox{!}{190px}{
\hspace*{-90pt}\input{Digrams.tikz}\architectureEncC
}
\caption{ESPNet-C}
\label{fig:archC}
\end{subfigure}
\hfill
\begin{subfigure}[b]{0.35\columnwidth}
\resizebox{!}{190px}{
\hspace*{-90pt}\input{Digrams.tikz}\architecture
}
\caption{ESPNet}
\label{fig:espnetFullArch}
\end{subfigure}
\caption{The path from ESPNet-A to ESPNet. Red and green color boxes represent the modules responsible for down-sampling and up-sampling operations, respectively. Spatial-level $l$ is indicated on the left of every module in (a). We denote each module as (\# input channels, \# output channels). Here, Conv-$n$ represents $n\times n$ convolution.}
\label{fig:espNetVariants}
\end{figure}

Different variants of ESPNet are shown in \Fig \ref{fig:espNetVariants}. The first variant, ESPNet-A (\Fig \ref{fig:archA}), is a standard network that takes an RGB image as an input and learns representations at different spatial levels\footnote{At each spatial level $l$, the spatial dimensions of the feature maps are the same. To learn representations at different spatial levels, a down-sampling operation is performed (see \Fig \ref{fig:archA}).} using the ESP module to produce a segmentation mask. The second variant, ESPNet-B (\Fig \ref{fig:archB}), improves the flow of information inside ESPNet-A by sharing the feature maps between the previous strided ESP module and the previous ESP module. The third variant, ESPNet-C (\Fig \ref{fig:archC}), reinforces the input image inside ESPNet-B to further improve the flow of information. These three variants produce outputs whose spatial dimensions are $\frac{1}{8}{th}$ of the input image. The fourth variant, ESPNet (\Fig \ref{fig:espnetFullArch}), adds a light weight decoder (built using a principle of \textit{reduce-upsample-merge}) to ESPNet-C that outputs the segmentation mask of the same spatial resolution as the input image. 

To build deeper computationally efficient networks for edge devices without changing the network topology,  a hyper-parameter $\alpha$  controls the depth of the network;  the ESP module is repeated $\alpha_l$ times at spatial level $l$. CNNs require more memory at higher spatial levels (at $l=0$ and $l=1$) because of the high spatial dimensions of feature maps at these levels. To be memory efficient, neither the ESP nor the convolutional modules are repeated at these spatial levels. The building block functions used to build the ESPNet (from ESPNet-A to ESPNet) are discussed in \textbf{Appendix \ref{sec:detailsBlocks}}. 

\noindent \textbf{Dataset:} We evaluated the ESPNet on the Cityscapes dataset \cite{cordts2016cityscapes}, an urban visual scene understanding dataset that consists of 2,975 training, 500 validation, and 1,525 test high-resolution images. The dataset was captured across 50 cities and in different seasons. The task is to segment an image into 19 classes belonging to 7 categories (e.g. person and rider classes belong to the same category \textit{human}). We evaluated our networks on the test set using the Cityscapes \textit{online} server.

To study the generalizability, we tested the ESPNet on an unseen dataset. We used the Mapillary dataset \cite{MVD2017} for this task because of its diversity. We mapped the annotations (65 classes) in the validation set (\# 2,000 images) to seven categories in the Cityscape dataset.
To further study the segmentation power of our model, we trained and tested the ESPNet on two other popular datasets from different domains. First, we used the widely known PASCAL VOC dataset \cite{everingham2010pascal} that has 1,464 training images, 1,448 validation images, and 1,456 test images. The task is to segment an image into 20 foreground classes. We evaluate our networks on the test set (comp6 category) using the PASCAL VOC \textit{online} server. Following the convention, we used additional images from \cite{hariharan2011semantic,lin2014microsoft}. Secondly, we used a breast biopsy whole slide image dataset \cite{mehta2017learning}, chosen because tissue structures in biomedical images vary in size and shape and because this dataset allowed us to check the potential of learning representations from a large receptive field. The dataset consists of 30 training images and 28 validation images, whose average size is $10,000 \times 12,000$, much larger than natural scene images. The task is to segment the images into 8 biological tissue labels; details are in \cite{mehta2017learning}.

\noindent \textbf{Performance evaluation metrics:} Most traditional CNNs measure network performance in terms of accuracy, latency, number of network parameters, and network size (e.g. \cite{howard2017mobilenets,zhang2017shufflenet,paszke2016enet,romera2018erfnet,iandola2016squeezenet}). These metrics provide high-level insight about the network, but fail to demonstrate the efficient usage of limited available hardware resources. In addition to these metrics, we introduce several \textit{system-level metrics} to characterize the performance of a CNN on resource-constrained devices \cite{yasin2014deep,wu2015performance}.

\noindent \textbf{\textit{Segmentation accuracy}} is measured as a mean Intersection over Union (mIOU) score between the ground truth and the predicted segmentation mask.

\noindent \textbf{\textit{Latency}} represents the amount of time a CNN network takes to process an image. This is usually measured in terms of frames per second (FPS).

\noindent \textbf{\textit{Network parameters}} represents the number of parameters learned by the network. 

\noindent \textbf{\textit{Network size}} represents the amount of storage space required to store the network parameters. An efficient network should have a smaller network size. 

\noindent \textbf{\textit{Sensitivity to GPU frequency}} measures the computational capability of an application and is defined as a ratio of percentage change in execution time to the percentage change in GPU frequency. A higher value indicates that the application tends to utilize the GPU more efficiently.

\noindent \textbf{\textit{Utilization rates}} measures the utilization of compute resources (CPU, GPU, and memory) while running on an edge device. In particular, computing units in edge devices (e.g. Jetson TX2) share memory between CPU and GPU.

\noindent \textbf{\textit{Warp execution efficiency}} is defined as the average percentage of active threads in each executed warp. GPUs schedule threads in the form of warps, and each thread inside the warp is executed in \textit{single instruction multiple data} fashion. A high value of warp execution efficiency represents efficient usage of GPU.

\noindent \textbf{\textit{Memory efficiency}} is the ratio of number of bytes requested/stored to the number of bytes transfered from/to device (or shared) memory to satisfy load/store requests. Since memory transactions are in blocks, this metric allows us to determine how efficiently we are using the memory bandwidth.

\noindent \textbf{\textit{Power consumption}} is the amount of average power consumed by the application during inference.

\noindent \textbf{Training details:} ESPNet networks were trained using PyTorch \cite{paszke2017automatic} with CUDA 9.0 and cuDNN back-ends. ADAM \cite{kingma2014adam} was used with an initial learning rate of $0.0005$, and decayed by two after every 100 epochs and with a weight decay of 0.0005. An inverse class probability weighting scheme was used in the cross-entropy loss function to address the class imbalance \cite{paszke2016enet,romera2018erfnet}. Following \cite{paszke2016enet,romera2018erfnet}, the weights were initialized randomly. Standard strategies, such as scaling, cropping and flipping, were used to augment the data. The image resolution in the Cityscape dataset is $2048 \times 1024$, and all the accuracy results were reported at this resolution. For training the networks, we sub-sampled the RGB images by two.  When the output resolution was smaller than $2048 \times 1024$, the output was up-sampled using bi-linear interpolation. For training on the PASCAL dataset, we used a fixed image size of $512 \times 512$. For the WSI dataset, the patch-wise training approach was followed \cite{mehta2017learning}. ESPNet was trained in two stages. First, ESPNet-C was trained with down-sampled annotations. Second, a light-weight decoder was attached to ESPNet-C and then, the entire ESPNet network was trained.

Three different GPU devices were used for our experiments: (1) a  desktop with a NVIDIA TitanX GPU (3,584 CUDA cores), (2) a laptop with a NVIDIA GTX-960M GPU (640 CUDA cores), and (3) an edge device with NVIDIA Jetson TX2 (256 CUDA cores). See \textbf{Appendix \ref{sec:hardwareDetailsSup}} for more details about the hardware. Unless and otherwise stated explicitly, statistics, such as power consumption and inference speed, are reported for an RGB image of size $1024 \times 512$ averaged over 200 trials. For collecting the hardware-level statistics, NVIDIA's and Intel's hardware profiling and tracing tools, such as NVPROF \cite{nvprof}, Tegrastats \cite{tegra}, and PowerTop \cite{powertop}, were used. In our  experiments, we will refer to ESPNet with $\alpha_2=2$ and $\alpha_3=8$ as ESPNet until and otherwise stated explicitly. 

\subsection{Results on the Cityscape dataset}
\label{ssec:sotaCity}

\begin{figure}[b!]
\centering
\begin{subfigure}[b]{0.48\columnwidth}
\centering
\includegraphics[height=90px]{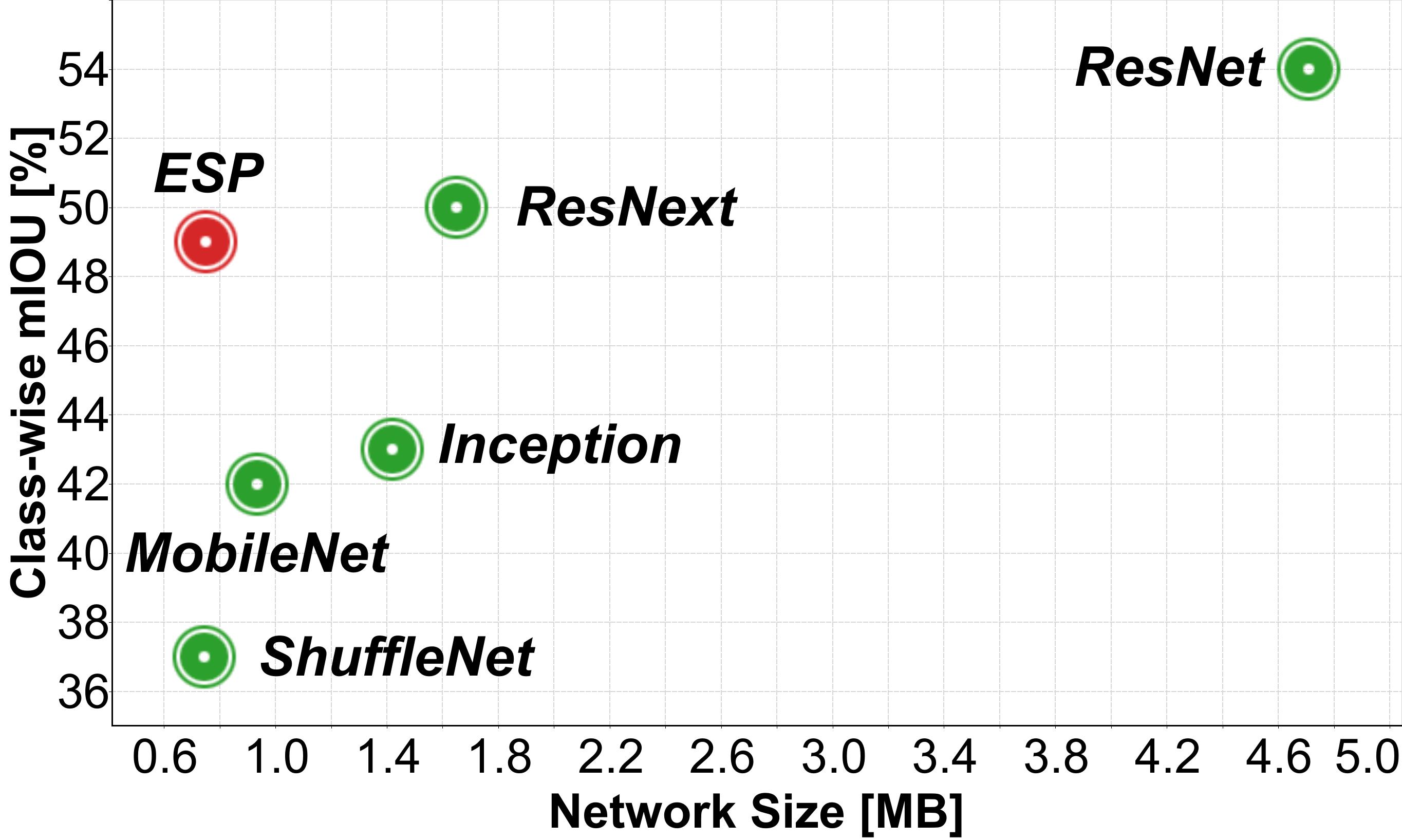}
\caption{Accuracy vs. network size}
\end{subfigure}
\hfill
\begin{subfigure}[b]{0.48\columnwidth}
\centering
\includegraphics[height=90px]{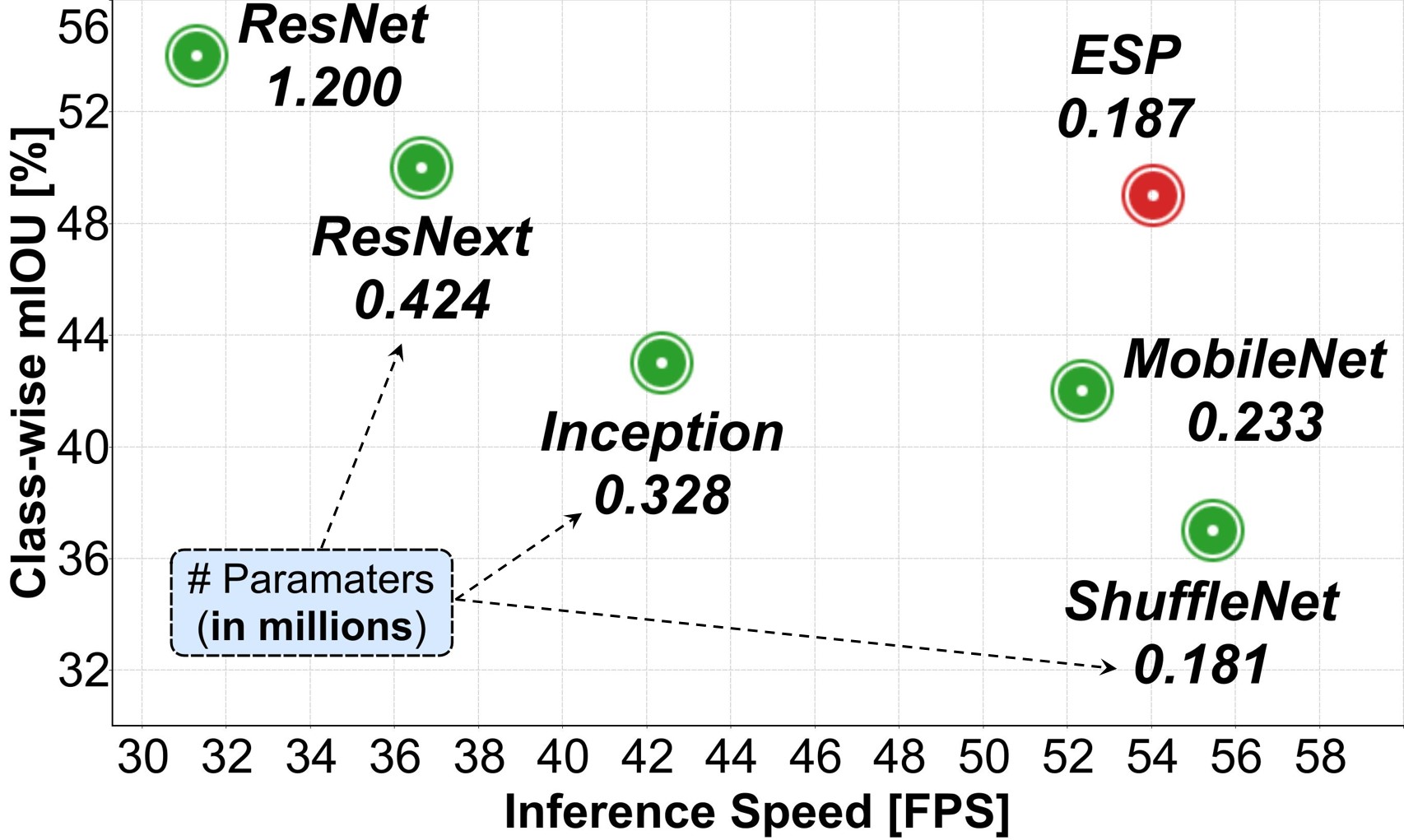}
\caption{Accuracy vs. speed (\textit{laptop})}
\end{subfigure}
\caption{Comparison between state-of-the-art efficient convolutional modules. For a fair comparison between different modules, we used $K=5$, $d=\frac{N}{K}$, $\alpha_2=2$, and $\alpha_3=3$. We used standard strided convolution for down-sampling. For ShuffleNet, we used $g=4$ and $K=4$ so that the resultant ESPNet-C network has the same complexity as with the ESP block.}
\label{fig:blockCompare}
\end{figure}

\noindent \textbf{Comparison with state-of-the-art efficient convolutional modules}: In order to understand the ESP module, we replaced the ESP modules in ESPNet-C with state-of-the-art efficient convolutional modules, sketched in \Fig \ref{fig:compareBlocks} (MobileNet \cite{howard2017mobilenets}, ShuffleNet \cite{zhang2017shufflenet}, Inception \cite{szegedy2015going,szegedy2016rethinking,SzegedyIV16InceptionV4}, ResNext \cite{xie2017aggregated}, and ResNet \cite{he2016deep}) and evaluate their  performance on the Cityscape validation dataset. We did not compare with ASP \cite{chen2016deeplab}, because it is computationally expensive and not suitable for edge devices. \Fig \ref{fig:blockCompare} compares the performance of ESPNet-C with different convolutional modules. Our ESP module outperformed  MobileNet and ShuffleNet modules by 7\% and 12\%, respectively, while learning a similar number of parameters and having comparable network size and inference speed. Furthermore, the ESP module delivered comparable accuracy to ResNext and Inception more efficiently. A basic ResNet module (stack of two $3\times 3$ convolutions with a skip-connection) delivered the best performance, but had to learn $6.5\times$ more parameters.

\begin{figure}[t!]
\centering
\begin{subfigure}[b]{0.24\columnwidth}
\centering
\resizebox{\columnwidth}{!}{
\begin{tabular}{|l|c|c|}
\cline{2-3}
\multicolumn{1}{l|}{} & \multicolumn{2}{c|}{\textbf{mIOU}}\\ 
\hline
\textbf{Network} & \textbf{Class} & \textbf{Category}\\ 
\hline
\textbf{ENet} \cite{paszke2016enet} & 58.3 & 80.4 \\ \hline
\textbf{ERFNet} \cite{romera2018erfnet} & 68.0 & 86.5 \\ \hline
\textbf{SQNet} \cite{jaderberg2014speeding} & 59.8 & 84.3\\ \hline
\textbf{SegNet} \cite{badrinarayanan2017segnet} & 57.0 & 79.1 \\ \hline
\textbf{ESPNet} (Ours) & 60.3 & 82.2 \\ 
\hline
\multicolumn{3}{c}{} \\
\hline
\textbf{FCN-8s} \cite{badrinarayanan2017segnet} & 65.3 & 85.7\\ \hline
\textbf{DeepLab-v2} \cite{chen2016deeplab} & 70.4 & 86.4 \\ \hline
\textbf{PSPNet} \cite{zhao2017pyramid} & \textbf{78.4} & \textbf{90.6}\\ \hline
\end{tabular}
}
\caption{Test set}
\label{tab:compareSt}
\end{subfigure}
\hfill
\begin{subfigure}[b]{0.36\columnwidth}
\centering
\includegraphics[width=\columnwidth]{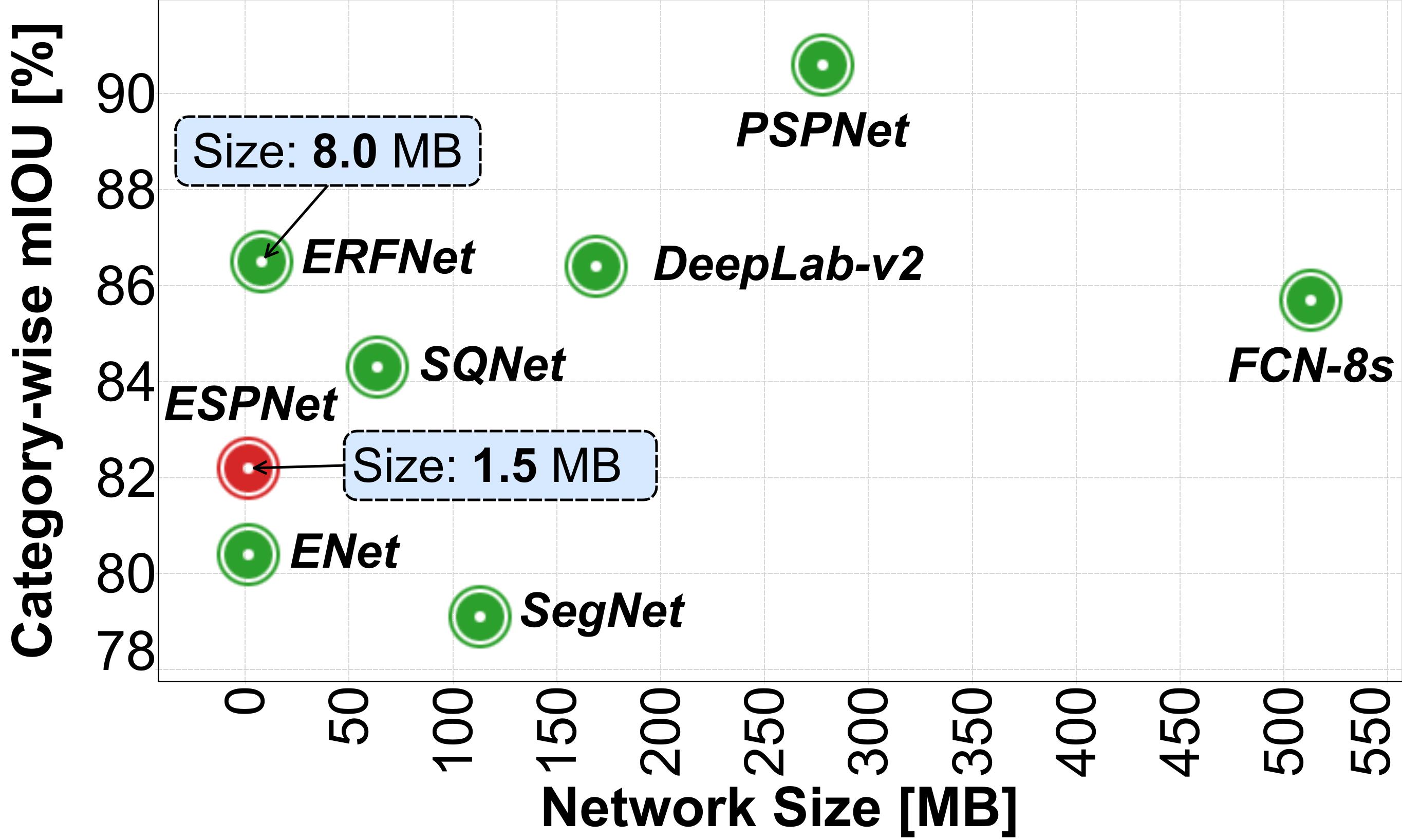}
\caption{Accuracy vs. network size}
\end{subfigure}
\hfill
\begin{subfigure}[b]{0.36\columnwidth}
\centering
\includegraphics[width=\columnwidth]{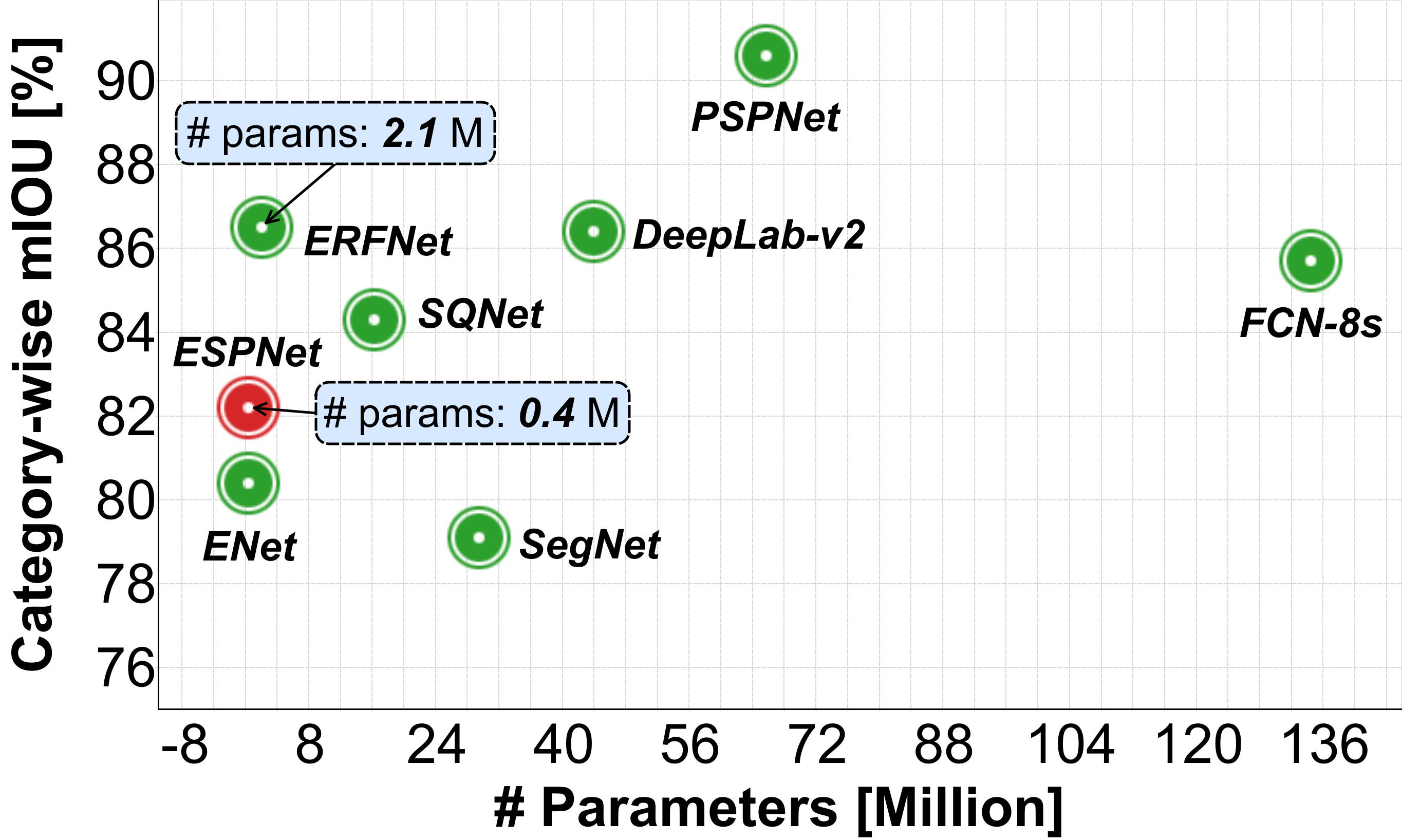}
\caption{Accuracy vs. \# parameters}
\end{subfigure}
\vspace{2mm}
\begin{subfigure}[b]{0.48\columnwidth}
\centering
\includegraphics[height=80px]{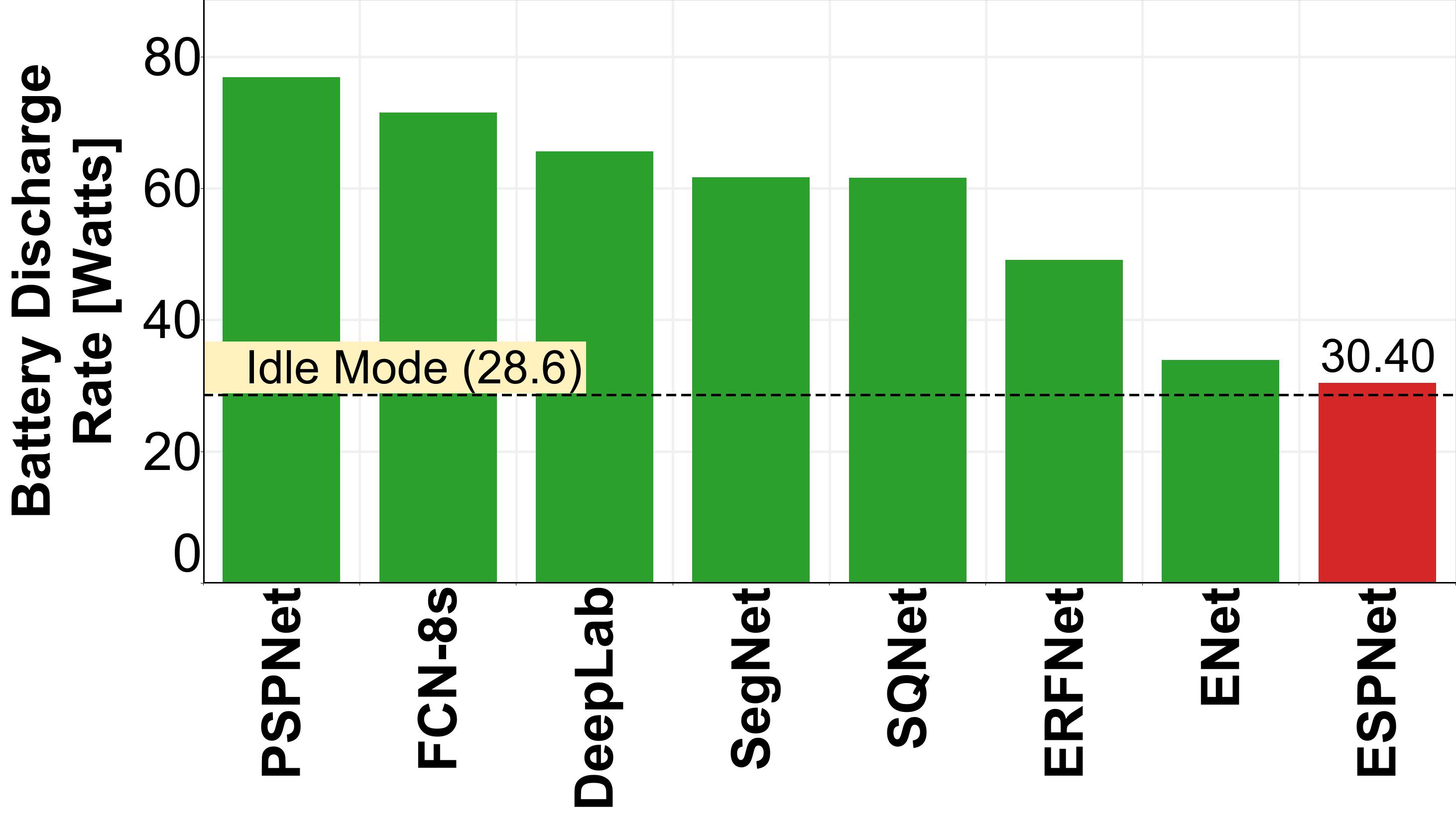}
\caption{Battery discharge rate vs. network (\textit{laptop})}
\label{fig:batDis}
\end{subfigure}
\hfill
\begin{subfigure}[b]{0.48\columnwidth}
\centering
\includegraphics[height=85px]{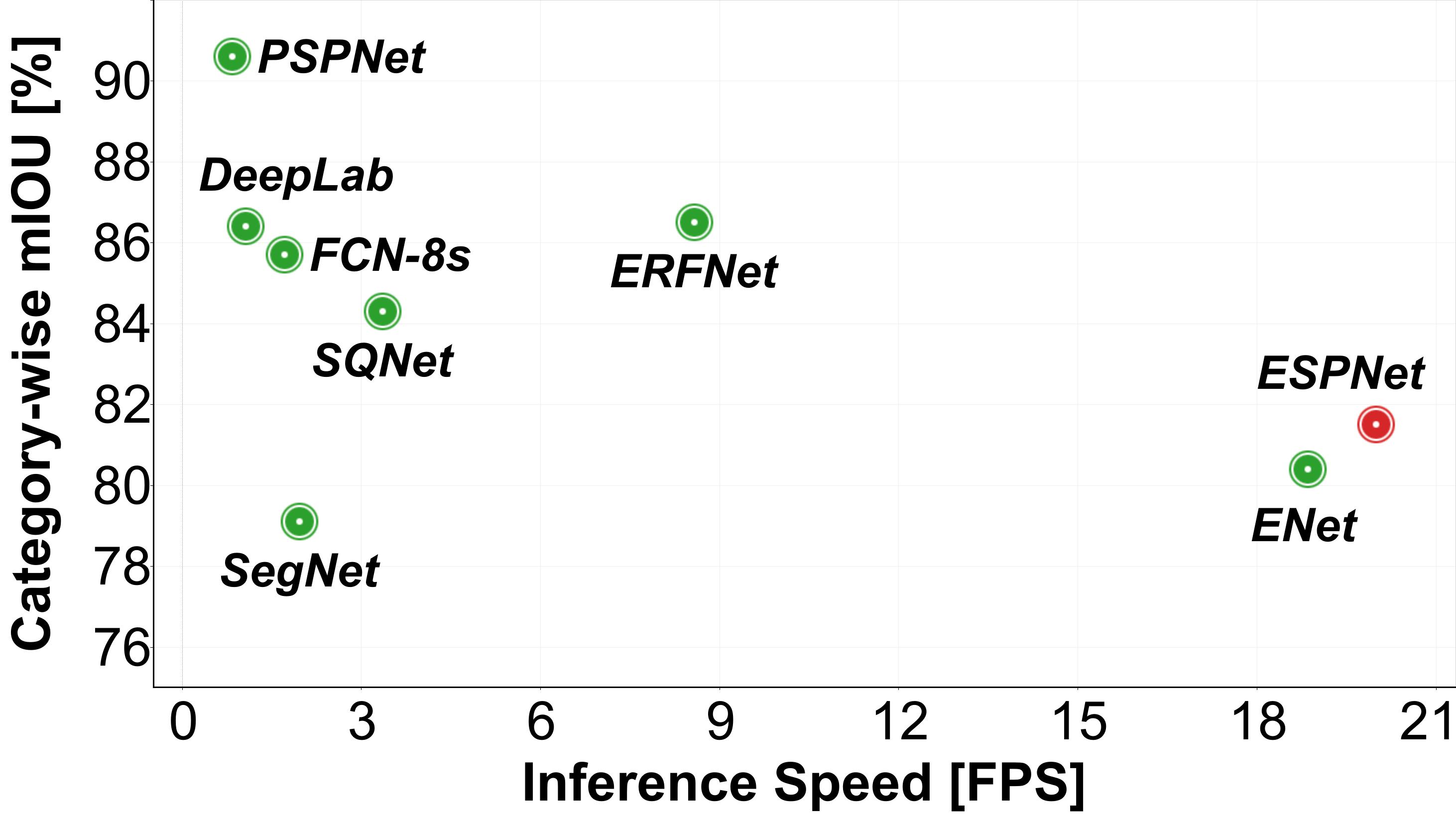}
\caption{Accuracy vs. speed (\textit{laptop})}
\label{fig:overAccComp}
\end{subfigure}
\vspace{2mm}
\begin{subfigure}[b]{0.48\columnwidth}
\centering
\includegraphics[height=85px]{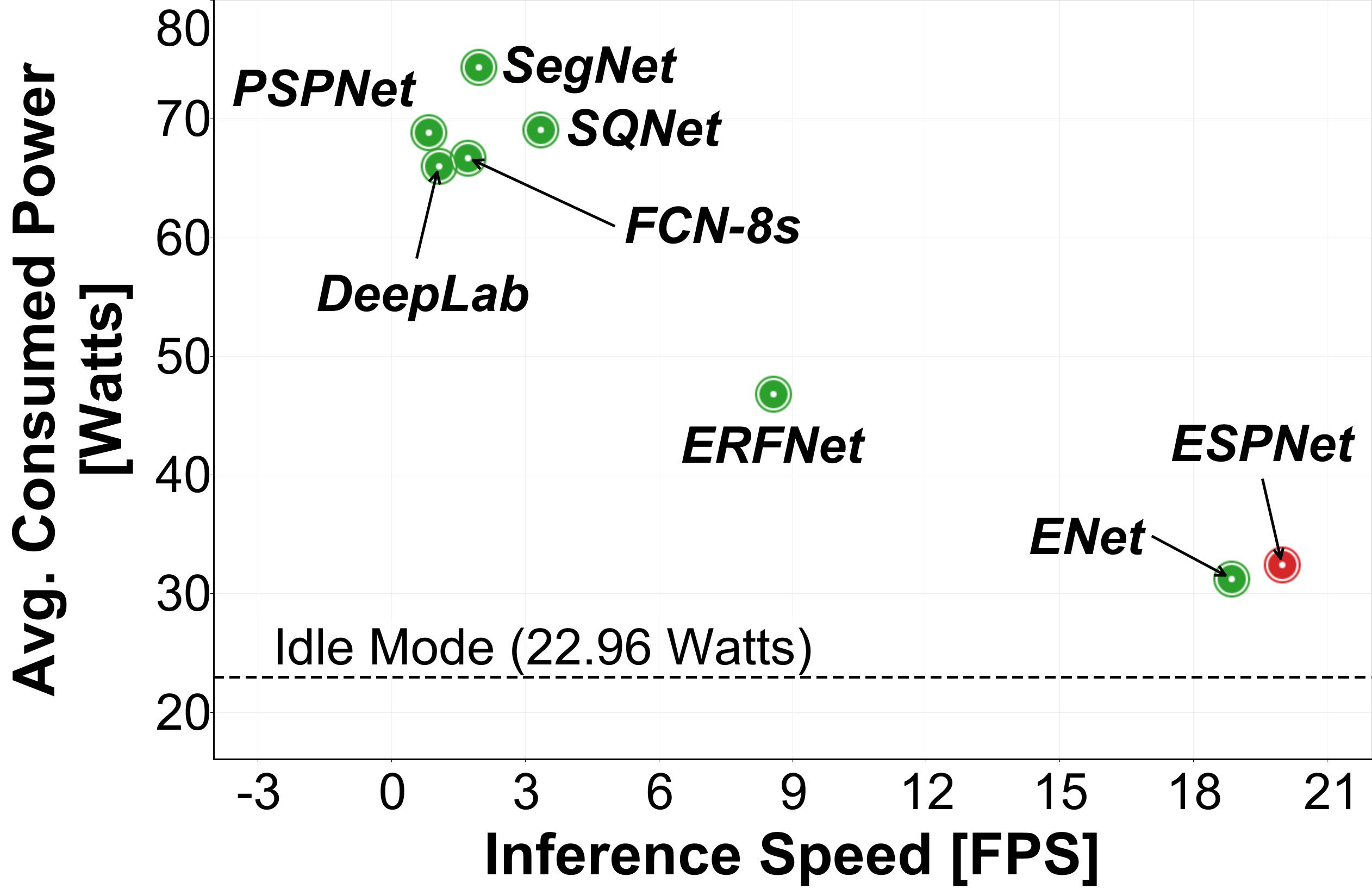}
\caption{Power consumption vs. speed (\textit{laptop})}
\label{fig:powerLap}
\end{subfigure}
\hfill
\begin{subfigure}[b]{0.5\columnwidth}
\centering
\includegraphics[height=85px]{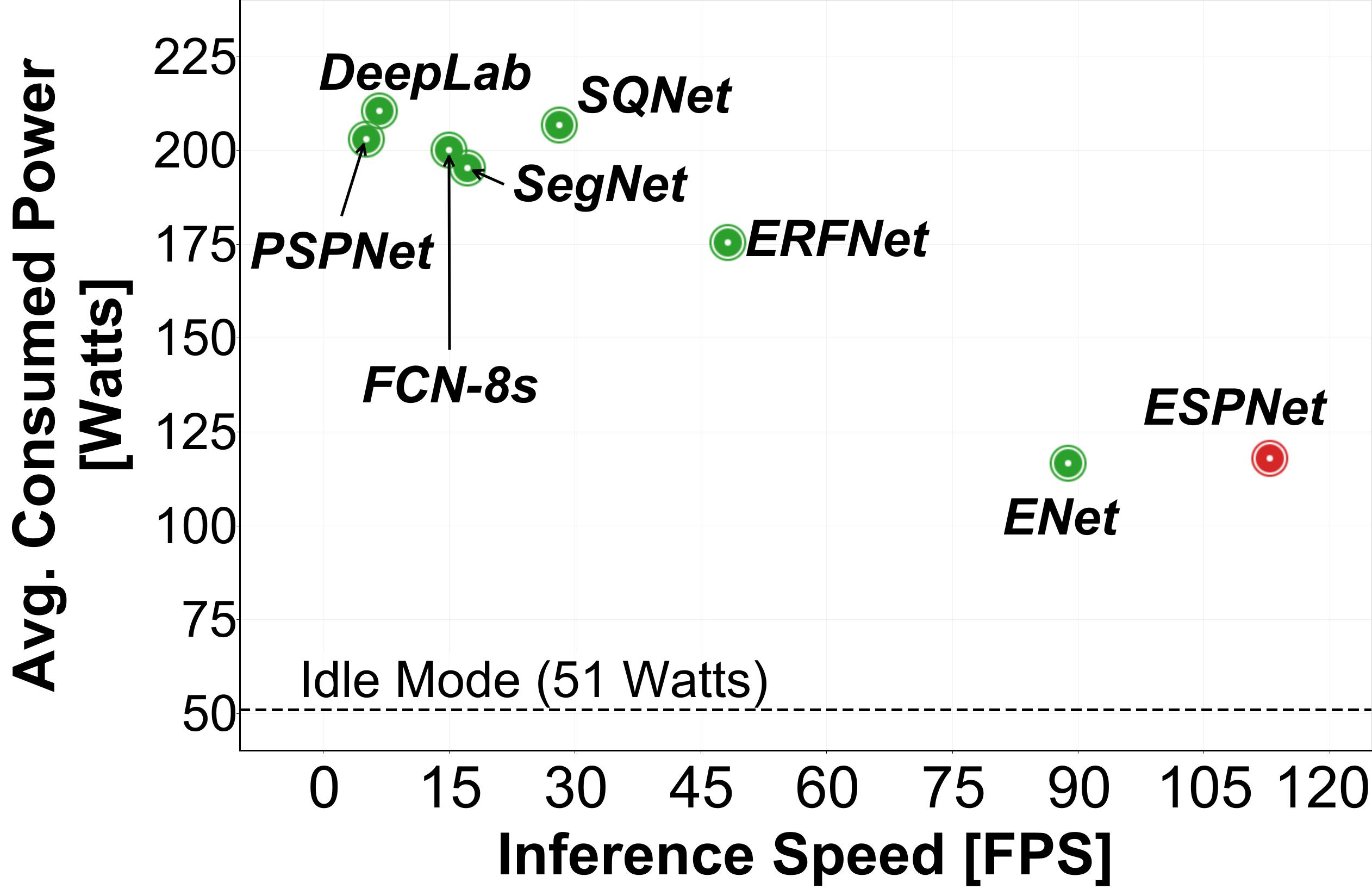}
\caption{Power consumption vs. speed (\textit{desktop})}
\label{fig:powerDesk}
\end{subfigure}
\caption{Comparison between state-of-the-art segmentation methods on the Cityscape test set on two different devices. All networks (FCN-8s \cite{long2015fully}, SegNet \cite{badrinarayanan2017segnet}, SQNet \cite{treml2016speeding}, ENet \cite{paszke2016enet}, DeepLab-v2 \cite{chen2016deeplab}, PSPNet \cite{zhao2017pyramid}, and ERFNet \cite{romera2018erfnet}) were without conditional random field and converted to PyTorch for a fair comparison. Best viewed in color.}
\label{fig:compareSt}
\end{figure}

\noindent \textbf{Comparison with state-of-the-art segmentation methods}: We compared the performance of ESPNet with state-of-the-art semantic segmentation networks. These networks either use a pre-trained network (\textit{VGG} \cite{simonyan2014very}: FCN-8s \cite{long2015fully} and SegNet \cite{badrinarayanan2017segnet}, \textit{ResNet} \cite{he2016deep}: DeepLab-v2 \cite{chen2016deeplab} and PSPNet \cite{zhao2017pyramid}, and \textit{SqueezeNet} \cite{iandola2016squeezenet}: SQNet \cite{treml2016speeding}) or were trained from scratch (ENet \cite{paszke2016enet} and ERFNet \cite{romera2018erfnet}). \Fig \ref{fig:compareSt} compares ESPNet with state-of-the-art methods. ESPNet is 2\% more accurate than ENet \cite{paszke2016enet}, while running $1.27\times$ and $1.16\times$ faster on a desktop and a laptop, respectively. ESPNet makes some mistakes between classes that belong to the same category, and hence has a lower class-wise accuracy (see \textbf{Appendix \ref{sec:resultsSup}} for the confusion matrix). For example, a rider can be confused with a person. However, ESPNet delivers a good category-wise accuracy. ESPNet had 8\% lower category-wise mIOU than PSPNet \cite{zhao2017pyramid}, while learning $180\times$ fewer parameters. ESPNet had lower power consumption, had lower battery discharge rate, and was significantly faster than state-of-the-art methods, while still achieving a competitive category-wise accuracy; this makes ESPNet suitable for segmentation on edge devices. ERFNet, an another efficient segmentation network, delivered good segmentation accuracy, but has $5.5\times$ more parameters, is $5.44 \times$ larger, consumes more power, and has a higher battery discharge rate than ESPNet. Also, ERFNet does not utilize limited available hardware resources efficiently on edge devices (Section \ref{ssec:perfEval}).

\subsection{Segmentation results on other datasets}
\label{ssec:mapillary}
\noindent \textbf{Unseen dataset:} Table \ref{tab:mapResults} compares the performance of ESPNet to that of ENet \cite{paszke2016enet} and ERFNet \cite{romera2018erfnet} on an unseen dataset. These networks were trained on the Cityscapes dataset \cite{cordts2016cityscapes} and tested on the Mapillary (unseen) dataset \cite{MVD2017}. ENet and ERFNet were chosen, because ENet was one of the most power efficient segmentation networks, while ERFNet has high accuracy and moderate efficiency. Our experiments show that ESPNet learns good generalizable representations of objects and outperforms ENet and ERFNet both qualitatively and quantitatively on the unseen dataset.

\noindent \textbf{PASCAL VOC 2012 dataset:} (Table \ref{tab:pasc}) On the PASCAL dataset, ESPNet is $4\%$ more accurate than SegNet, one of the smallest network on the PASCAL VOC, while learning $81\times$ fewer parameters. ESPNet is $22\%$ less accurate than PSPNet (one of the most accurate network on the PASCAL VOC) while learning $180\times$ fewer parameters.

\noindent \textbf{Breast biopsy dataset:} (Table \ref{tab:bbvs}) On the breast biopsy  dataset, ESPNet achieved the same accuracy as \cite{mehta2017learning} while learning $9.5\times$ less parameters.

\begin{table}[t!]
\centering
\begin{subtable}[b]{0.33\columnwidth}
\centering
\resizebox{0.8\columnwidth}{!}{
\begin{tabular}{l|c|c|}
\cline{2-3}
& \textbf{mIOU} & \textbf{\# Params} \\ 
\hline
\multicolumn{1}{|l|}{\textbf{ENet} \cite{paszke2016enet}} &  0.33 & 0.364\\ 
\hline
\multicolumn{1}{|l|}{\textbf{ERFNet} \cite{romera2018erfnet}}  &  0.25 & 2.06 \\ 
\hline
\multicolumn{1}{|l|}{\textbf{ESPNet}} & \textbf{0.40} & 0.364 \\ 
\hline
\end{tabular}
}
\caption{Mapillary validation set \cite{MVD2017}}
\label{tab:mapResults}
\end{subtable}
\hfill
\begin{subtable}[b]{0.6\columnwidth}
\begin{subfigure}[b]{\columnwidth}
\includegraphics[width=\columnwidth]{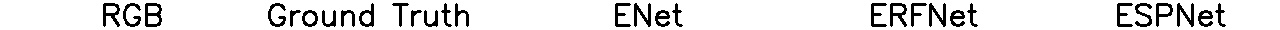}
\end{subfigure}
\begin{subfigure}[b]{\columnwidth}
\includegraphics[width=\columnwidth]{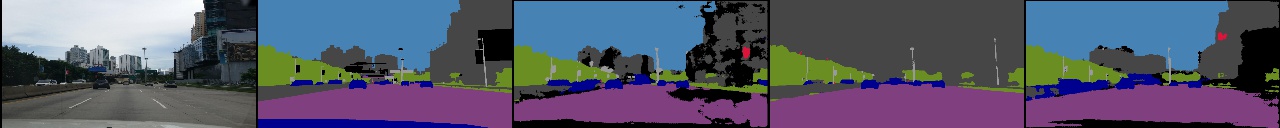}
\end{subfigure}
\begin{subfigure}[b]{\columnwidth}
\includegraphics[width=\columnwidth]{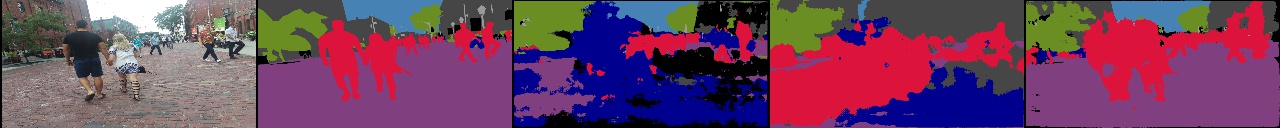}
\end{subfigure}
\caption{Mapillary validation set \cite{MVD2017} (unseen)}
\label{tab:mapResultsVis}
\end{subtable}
\vfill
\begin{subtable}[b]{0.6\columnwidth}
\centering
\resizebox{\columnwidth}{!}{
\begin{tabular}[b]{l|cccccccc}
\toprule
\textbf{Model} & \multicolumn{1}{c}{\textbf{ESPNet}} & \multicolumn{1}{c}{\textbf{SegNet}} & \multicolumn{1}{c}{\textbf{RefineNet}} & \multicolumn{1}{c}{\textbf{DeepLab}} & \multicolumn{1}{c}{\textbf{PSPNet}} & \multicolumn{1}{c}{\textbf{LRR}} & \multicolumn{1}{c}{\textbf{Dilation-8}} & \multicolumn{1}{c}{\textbf{FCN-8s}} \\
 & \multicolumn{1}{c}{\textbf{(Ours)}} & \multicolumn{1}{c}{\textbf{\cite{badrinarayanan2017segnet}}} & \multicolumn{1}{c}{\textbf{\cite{lin2017refinenet}}} & \multicolumn{1}{c}{\textbf{\cite{chen2016deeplab}}} & \multicolumn{1}{c}{\textbf{\cite{zhao2017pyramid}}} & \multicolumn{1}{c}{\textbf{\cite{ghiasi2016laplacian}}} & \multicolumn{1}{c}{\textbf{\cite{yu2015multi}}} & \multicolumn{1}{c}{\textbf{\cite{long2015fully}}} \\
 \midrule
\textbf{\# Params} & \textbf{0.364} & 29.5 & 42.6 & 44.04 & 65.7 & 48 & 141.13 & 134.5 \\
\textbf{mIOU} & 63.01 & 59.10 & 82.40 & 79.70 & \textbf{85.40} & 79.30 & 75.30 & 67.20 \\
\bottomrule
\end{tabular}
}
\caption{PASCAL VOC test set \cite{everingham2010pascal}}
\label{tab:pasc}
\end{subtable}
\hfill
\begin{subtable}[b]{0.38\columnwidth}
\centering
\resizebox{0.8\columnwidth}{!}{
\begin{tabular}{l|c|cc}
\toprule 
\textbf{Model} & \textbf{Module} & \textbf{mIOU} & \textbf{\# Params} \\
\midrule
ESPNet (Ours)$^\star$ & ESP & 44.03 & \textbf{2.75} \\
SegNet \cite{badrinarayanan2017segnet} & VGG & 37.6 & 12.80 \\
Mehta \etal \cite{mehta2017learning} & ResNet & \textbf{44.20} & 26.03 \\
\bottomrule
\end{tabular}
}
\caption{Breast biopsy validation set \cite{mehta2017learning}}
\label{tab:bbvs}
\end{subtable}
\caption{Results on different datasets. Here, the number of parameters are in million. $^\star$ For more details, please see \cite{mehta2018ynet}. See \textbf{Appendix \ref{sec:resultsSup}} for more qualitative results.}
\label{tab:pascal}
\end{table}

\subsection{Performance analysis on an edge device}
\label{ssec:perfEval}
\begin{figure}[t!]
	\centering
	\begin{subfigure}[b]{0.16\columnwidth}
		\centering
		\resizebox{\columnwidth}{!}{
			\begin{tabular}{|l|c|}
				\hline
				\textbf{Network} & \textbf{Size}\\
				\hline
				\textbf{ENet} & 1.64 MB\\
				\hline
				\textbf{ERFNet} & 7.95 MB\\
				\hline
				\textbf{ESPNet} & \textbf{1.46 MB} \\
				\hline
			\end{tabular}
		}
		\caption{}
		\label{fig:modelSize}
	\end{subfigure}
	\hfill
	\begin{subfigure}[b]{0.25\columnwidth}
		\centering
		\includegraphics[height=50px]{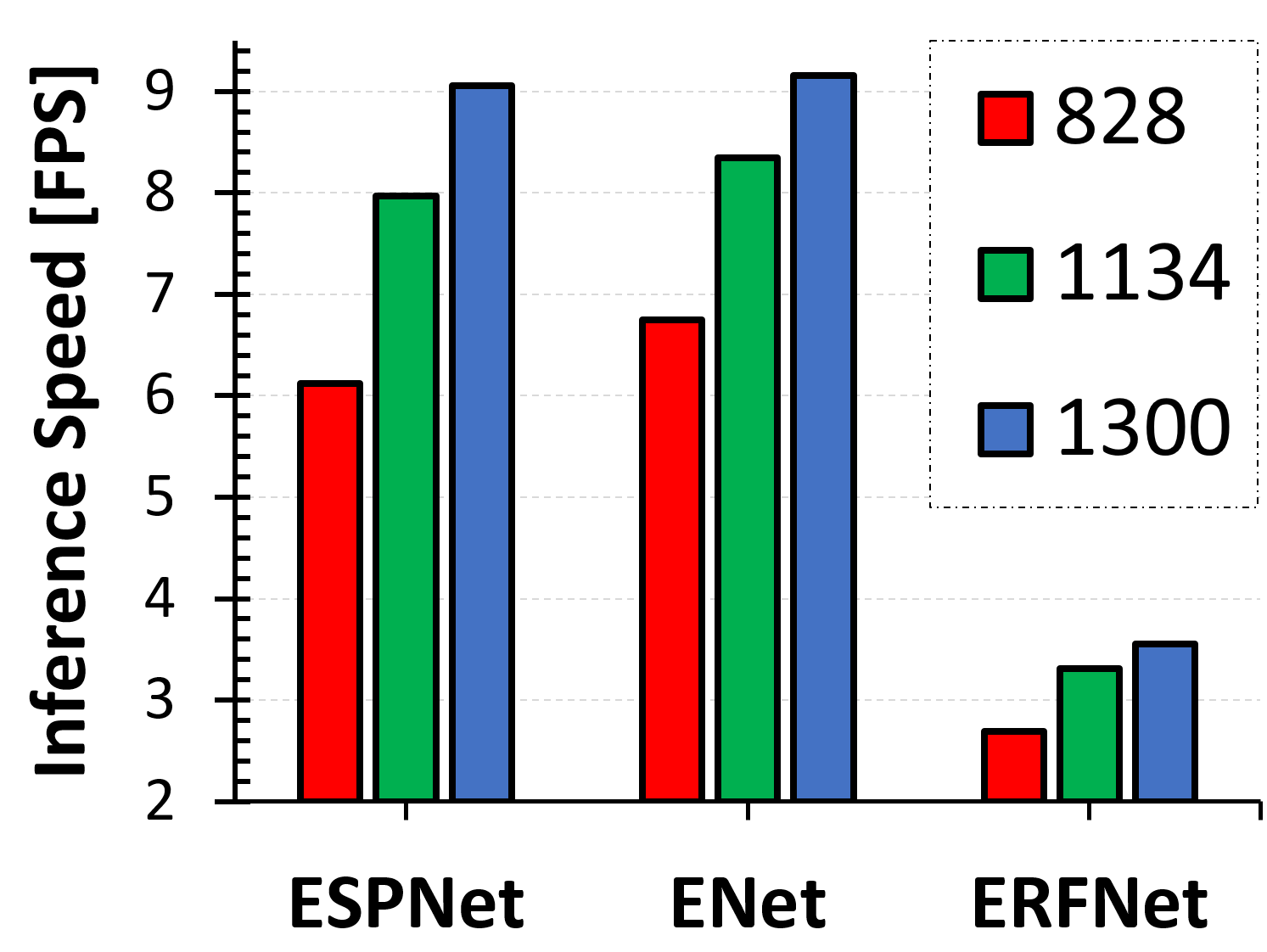}
		\caption{}
		\label{fig:infSpeed}
	\end{subfigure}
	\hfill
	\begin{subfigure}[b]{0.27\columnwidth}
		\centering
		\resizebox{\columnwidth}{!}{
			\begin{tabular}{|ll|c|c|}
				\hline
				\multirow{2}{*}{\textbf{Network}} &  & \multicolumn{2}{l|}{\textbf{Sensitivity to GPU freq.}} \\ \cline{3-4} 
				&  & \multicolumn{1}{l|}{\textbf{828 to 1134}} & \multicolumn{1}{l|}{\textbf{1134 to 1300}} \\ \hline
				\multicolumn{2}{|l|}{\textbf{ENet}} & 71\% & 70\% \\ 
				\hline
				\multicolumn{2}{|l|}{\textbf{ERFNet}} & 69\% & 53\% \\ 
				\hline
				\multicolumn{2}{|l|}{\textbf{ESPNet}}& 86\% & 95\% \\ 
				\hline
			\end{tabular}
		}
		\caption{}
		\label{fig:sensitivity}
	\end{subfigure}
	\hfill
	\begin{subfigure}[b]{0.24\columnwidth}
		\centering
		\resizebox{\columnwidth}{!}{
			\begin{tabular}{|l|c|c|c|}
				\hline
				\multirow{2}{*}{\textbf{Network}} & \multicolumn{3}{c|}{\textbf{Utilization (\%)}} \\ \cline{2-4} 
				& \textbf{CPU} & \textbf{GPU} & \textbf{Memory} \\ \hline
				\multicolumn{1}{|l|}{\textbf{ENet}} & 20.5 & 99.00 & 50.6 \\ \hline
				\multicolumn{1}{|l|}{\textbf{ERFNet}} & \textbf{19.7} & 99.00 & 61.3 \\ \hline
				\multicolumn{1}{|l|}{\textbf{ESPNet}} & 20.3 & 99.00 & \textbf{44.0} \\ \hline
			\end{tabular}
		}
		\caption{}
		\label{fig:utilization}
	\end{subfigure}
	\vfill
	\begin{subfigure}[b]{0.32\columnwidth}
		\includegraphics[height=60px]{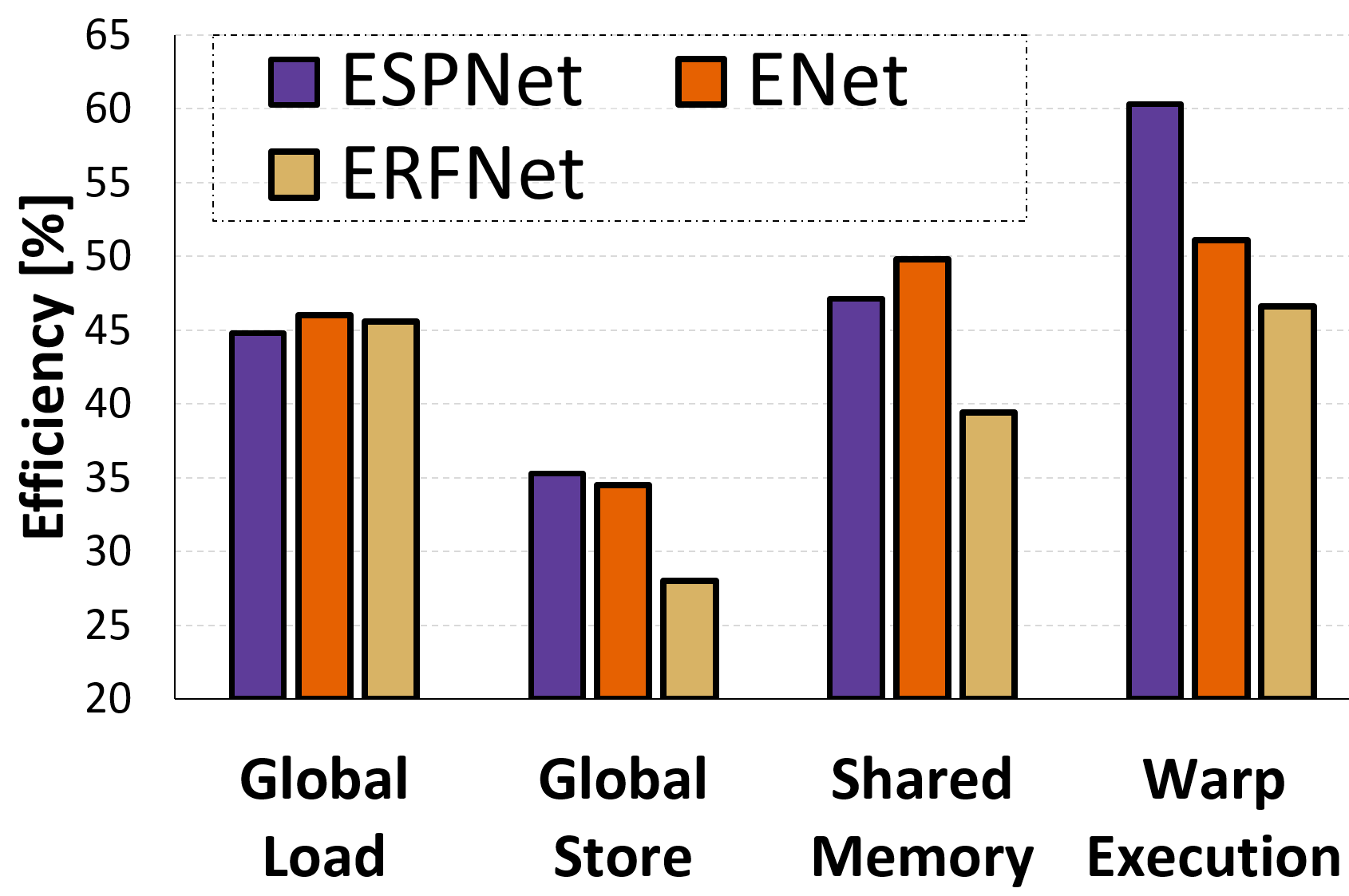}
		\caption{}
		\label{fig:efficiency}
	\end{subfigure}
	\hfill
	\begin{subfigure}[b]{0.32\columnwidth}
		\includegraphics[height=60px]{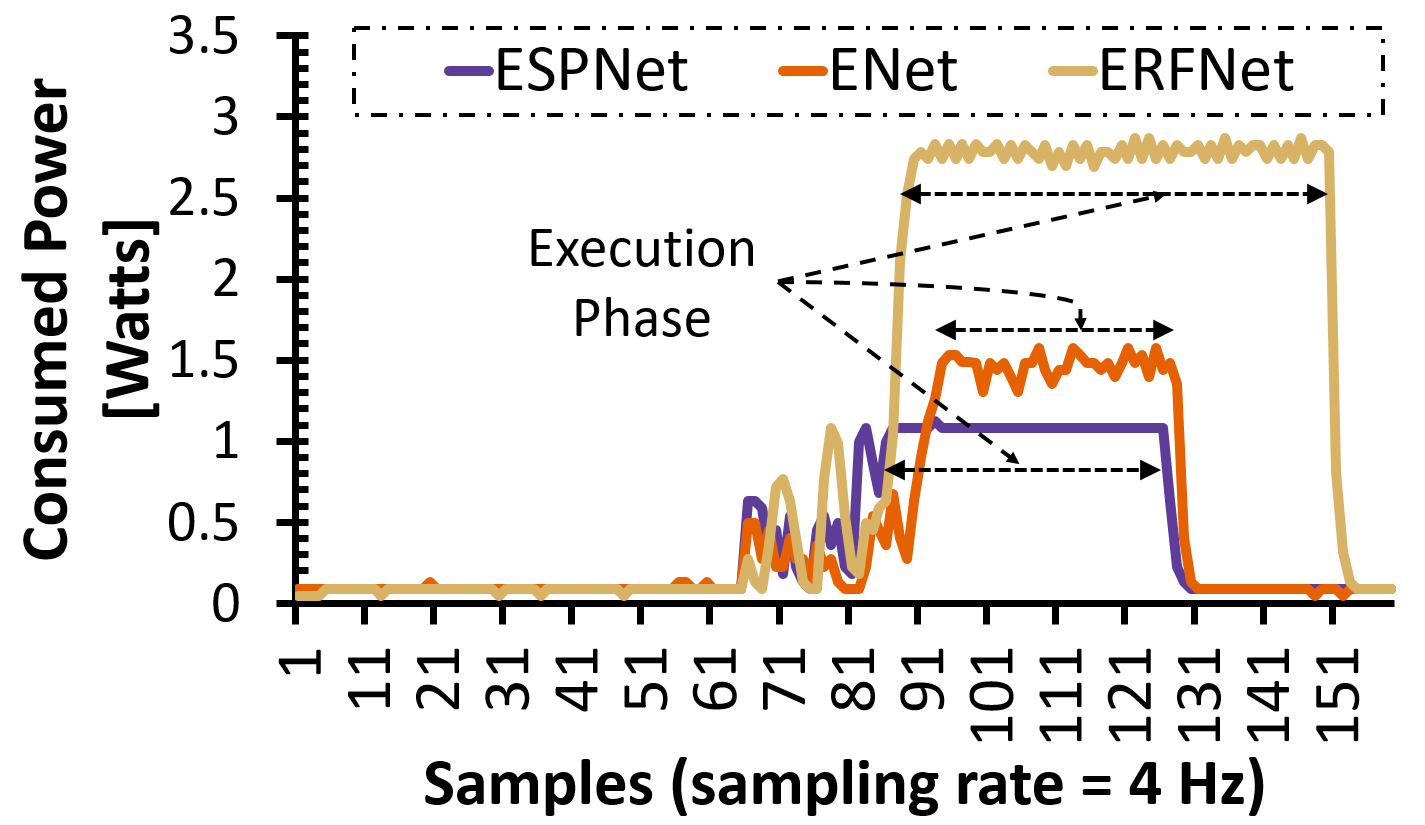}
		\caption{GPU freq. @ 828 MHz}
		\label{fig:maxQ}
	\end{subfigure}
	\hfill
	\begin{subfigure}[b]{0.32\columnwidth}
		\includegraphics[height=60px]{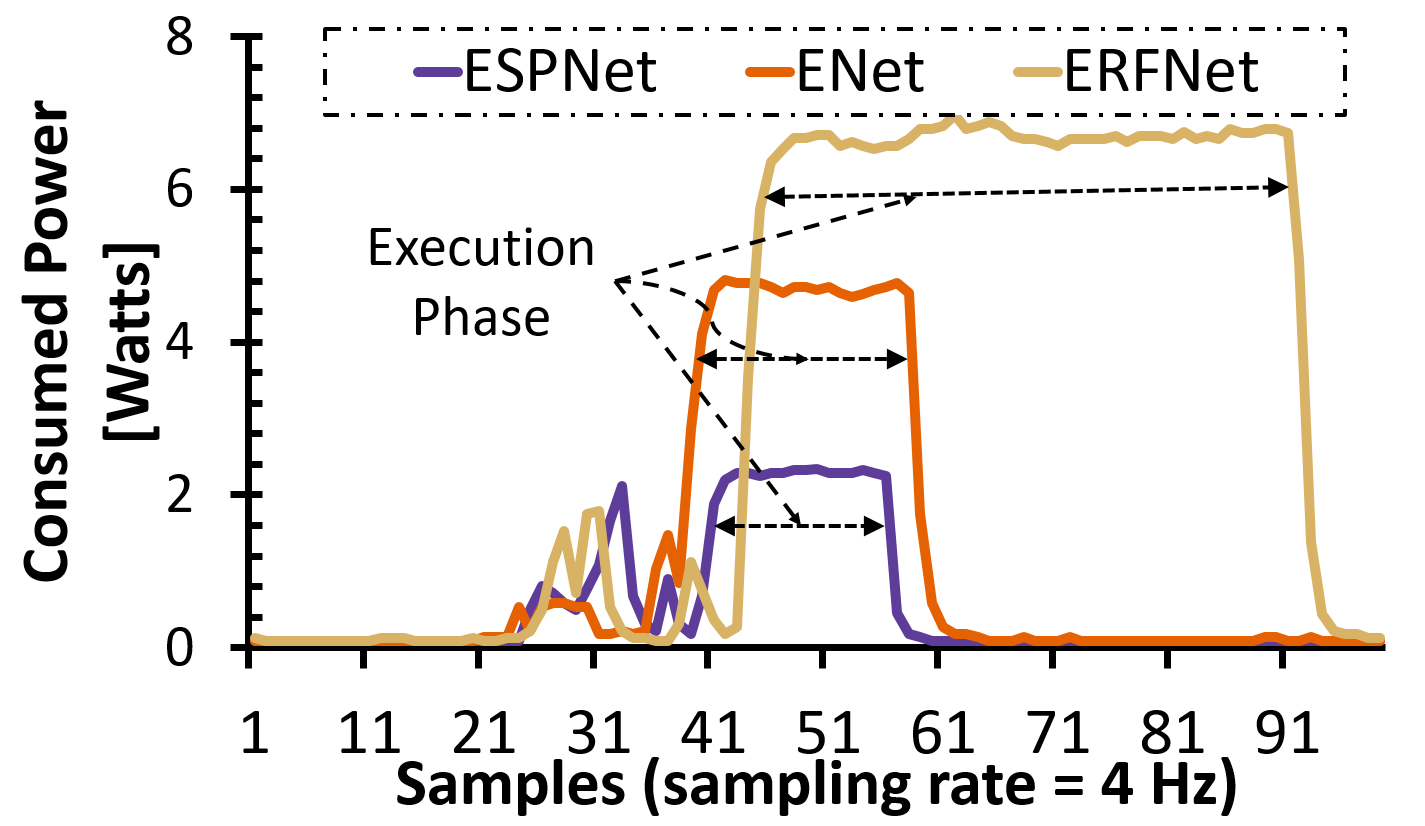}
		\caption{GPU freq. @ 1,134 MHz}
		\label{fig:maxP}
	\end{subfigure}
	\caption{Performance analysis of ESPNet with ENet and ERFNet on a NVIDIA Jetson TX2: (a) network size, (b) inference speed vs. GPU frequency (in MHz), (c) sensitivity analysis, (d) utilization rates,  (e) efficiency rates, and (f, g) power consumption at two different GPU frequencies. In (d), the statistics for the network's initialization phase were not considered, because they were the same across all networks. See \textbf{Appendix  \ref{sec:resourceUtilTop}} for time vs. utilization plots. Best viewed in color.}
	\label{fig:perfCharacter}
\end{figure}

We measure the performance on the NVIDIA Jetson TX2, a computing platform for edge devices. Performance analysis results are given in \Fig \ref{fig:perfCharacter}. 

\noindent \textbf{Network size}: \Fig \ref{fig:modelSize} compares the uncompressed 32-bit network size of ESPNet with ENet and ERFNet. ESPNet had a $1.12\times$ and $5.45\times$ smaller network than ENet and ERFNet, respectively, which reflects well on  the architectural design of ESPNet. 

\noindent \textbf{Inference speed and sensitivity to GPU frequency}: \Fig \ref{fig:infSpeed} compares the inference speed of ESPNet with ENet and ERFNet. ESPNet had almost the same frame rate as ENet, but it was more sensitive to GPU frequency (\Fig \ref{fig:sensitivity}). As a consequence, ESPNet achieved a higher frame rate than ENet on high-end graphic cards, such as the GTX-960M and TitanX (see \Fig \ref{fig:compareSt}). For example, ESPNet is $1.27\times$ faster than ENet on an NVIDIA TitanX. ESPNet is about $3\times$ faster than ERFNet on an NVIDIA Jetson TX2.

\noindent \textbf{Utilization rates}: \Fig \ref{fig:utilization} compares the CPU, GPU, and memory utilization rates of different networks. These networks are throughput intensive, and therefore, GPU utilization rates are high, while CPU utilization rates are low for these networks. Memory utilization rates are significantly different for these networks. The memory footprint of ESPNet is low in comparison to ENet and ERFNet, suggesting that ESPNet is suitable for memory-constrained devices.

\noindent \textbf{Warp execution efficiency}: \Fig \ref{fig:efficiency} compares the warp execution efficiency of ESPNet with ENet and ERFNet. The warp execution of ESPNet was about 9\% higher than ENet and about 14\% higher than ERFNet. This indicates that ESPNet has less warp divergence and promotes the efficient usage of limited GPU resources available on edge devices.  We note that warp execution efficiency gives a better insight into the utilization of GPU resources than the GPU utilization rate. GPU frequency will be busy even if few warps are active, resulting in a high GPU utilization rate.

\noindent \textbf{Memory efficiency}: (\Fig \ref{fig:efficiency}) All networks have similar global load efficiency, but ERFNet has a poor store and shared memory efficiency. This is likely due to the fact that ERFNet spends 20\% of the compute power performing memory alignment operations, while ESPNet and ENet spend 4.2\% and 6.6\% time for this operation, respectively. See \textbf{Appendix \ref{sec:topKernSUp}} for the compute-wise break down of different kernels.

\noindent \textbf{Power consumption}: \Fig \ref{fig:maxQ} and \ref{fig:maxP} compares the power consumption of ESPNet with ENet and ERFNet at two different GPU frequencies. The average power consumption (during network execution phase) of ESPNet, ENet, and ERFNet were 1 W, 1.5 W, and 2.9 W at a GPU frequency of 824 MHz and 2.2 W, 4.6 W, and 6.7 W at a GPU frequency of  1,134 MHz, respectively; suggesting ESPNet is a power-efficient network.

\subsection{Ablation studies on the Cityscapes: The path from ESPNet-A to ESPNet}
\label{ssec:ablationStudies}

Larger networks or ensembling the output of multiple networks delivers better performance \cite{zhao2017pyramid,chen2016deeplab,yu2017dilated}, but with ESPNet (sketched in \Fig \ref{fig:espNetVariants}), the goal is an efficient network for edge devices. To improve the performance of ESPNet while maintaining efficiency, a systematic study of design choices was performed. Table \ref{tab:ablationStudies} summarizes the results.

\begin{table}[t!]
	\centering
	\begin{subtable}[b]{0.18\columnwidth}
		\centering
		\resizebox{\columnwidth}{!}{
			\begin{tabular}{l|c|c}
				\toprule
				& \textbf{mIOU} & \textbf{\# Params}$^{\circ}$ \\
				\midrule
				ReLU & 0.36 & 0.183 \\
				PReLU & \textbf{0.38} & 0.183 \\
				\bottomrule
			\end{tabular}
		}
		\caption{}
		\label{tab:actiImp}
	\end{subtable}
	\hfill
	\begin{subtable}[b]{0.18\columnwidth}
		\centering
		\resizebox{\columnwidth}{!}{
			\begin{tabular}{l|c|c}
				\toprule
				\textbf{Module} & \textbf{mIOU} & \textbf{\# Params}$^{\circ}$ \\
				\midrule
				ESP & \textbf{0.39} & 0.183 \\
				\quad -RL & 0.37 & 0.183 \\
				\bottomrule
				\multicolumn{3}{l}{RL - residual learning}
			\end{tabular}
		}
		\caption{}
		\label{tab:blockImp}
	\end{subtable}
	\hfill
	\begin{subtable}[b]{0.22\columnwidth}
		\centering
		\resizebox{\columnwidth}{!}{
			\begin{tabular}{l|c|c}
				\toprule
				\textbf{Downsample} & \textbf{mIOU} & \textbf{\# Params}$^{\circ}$ \\
				\midrule
				Strided conv. & 0.38 & 0.274\\
				Strided ESP & \textbf{0.39} & \textbf{0.183} \\
				\bottomrule
			\end{tabular}
		}
		\caption{}
		\label{tab:downImp}
	\end{subtable}
	\hfill
	\begin{subtable}[b]{0.36\columnwidth}
		\centering
		\resizebox{\columnwidth}{!}{
			\begin{tabular}{c|c|c|c|c|c|}
				\toprule
				\textbf{ESPNet-C} & \multicolumn{2}{c|}{\textbf{ESP operations}} & \# \textbf{params} & \textbf{Network}   & \\
				\cline{2-3}
				\textbf{configuration} &  \textbf{Reduce} & \textbf{Transform} &  & \textbf{size} & \textbf{mIOU}\\
				\midrule
				C1 - ($\alpha_3 = 3$) & $3\times 3$ & SPC  & 0.276 & 1.2 MB & 50.8 \\
				C2 - ($\alpha_3 = 3$) & $1\times 1$ & SPC  & 0.187 & 0.8 MB & 49.0 \\
				C3 - ($\alpha_3 = 3$) & $1\times 1$ & SPC-s & 0.187 & 0.8 MB & 47.4 \\
				\bottomrule
			\end{tabular}
		}
		\caption{}
		\label{tab:effect}
	\end{subtable}
	\vfill
	\begin{subtable}[b]{0.36\columnwidth}
		\centering
		\resizebox{\columnwidth}{!}{
			\begin{tabular}{l|c|c|c|c|c|c}
				\toprule
				& \multicolumn{6}{c}{\textbf{Width divider $K$}} \\ \cline{2-7} 
				& \textbf{2} & \textbf{4} & \textbf{5} & \textbf{6} & \textbf{7} & \textbf{8} \\
				\midrule
				\multicolumn{1}{l|}{\textbf{mIOU}} & \textbf{0.415} & 0.378 & 0.381 & 0.359 & 0.321 & 0.303 \\ 
				\midrule
				\multicolumn{1}{l|}{\textbf{\# Params$^{\circ}$}} & 0.358 & 0.215 & 0.183 & 0.165 & 0.152& \textbf{0.143}\\ 
				\midrule
				\multicolumn{1}{l|}{\textbf{ERF} $(n^2 = n \times n)$} & $5^2$ & $17^2$ & $33^2$ & $65^2$ & $129^2$ & $\mathbf{257}^2$ \\
				\bottomrule
			\end{tabular}
		}
		\caption{}
		\label{tab:parallelImp}
	\end{subtable}
	\hfill
	\begin{subtable}[b]{0.2\columnwidth}
		\centering
		\resizebox{\columnwidth}{!}{
			\begin{tabular}{l|c|c}
				\toprule
				\textbf{Network} & \textbf{mIOU} & \textbf{\# Params}$^{\circ}$ \\
				\midrule
				ESPNet-A$^\star$ & 0.39 & \textbf{0.183} \\
				ESPNet-B & 0.40 & 0.186 \\
				ESPNet-C & \textbf{0.42} & 0.187\\
				ESPNet-C$^{\dagger}$ & \textbf{0.42} & 0.206\\
				\bottomrule
			\end{tabular}
		}
		\caption{}
		\label{tab:modelImp}
	\end{subtable}
	\hfill
	\begin{subtable}[b]{0.4\columnwidth}
		\resizebox{\columnwidth}{!}{
			\begin{tabular}{c|c|c|c||c|c|c}
				\toprule
				\multirow{3}{*}{$\alpha_3$}& \multicolumn{3}{|c||}{\textbf{ESPNet-C} (\Fig \ref{fig:archC})} & \multicolumn{3}{c}{\textbf{ESPNet} (\Fig \ref{fig:espnetFullArch})}\\
				\cline{2-7}
				& \multirow{2}{*}{\textbf{mIOU}} & \# \textbf{Params} & \textbf{Network} & \multirow{2}{*}{\textbf{mIOU}} & \# \textbf{Params} & \textbf{Network}\\
				&  & \textbf{(in million)} & \textbf{size} &  & \textbf{(in million)} & \textbf{size}\\
				\midrule
				3 & 49.0 & \textbf{0.187} & \textbf{0.75 MB} & 56.3 & \textbf{0.202} & \textbf{0.82 MB} \\
				5 & 51.2 & 0.252 & 1.01 MB & 57.9  & 0.267 & 1.07 MB \\
				8 & \textbf{53.3} & 0.349 & 1.40 MB & \textbf{61.4} &  0.364 & 1.46 MB\\
				\bottomrule
			\end{tabular}
		}
		\caption{}
		\label{tab:espnetCompare}
	\end{subtable}
	\caption{The path from ESPNet-A to ESPNet. Here, ERF represents effective receptive field, $^\star$ denotes that strided ESP was used for down-sampling, $^{\dagger}$ indicates that the input reinforcement method was replaced with input-aware fusion method \cite{mehta2017learning}, 
		and $^{\circ}$ denotes the values are in million. All networks in (a-c,e-f) are trained for 100 epochs, while networks in (d,g) are trained for 300 epochs. Here, SPC-s denotes that $3\times3$ standard convolutions are used instead of dilated convolutions in the spatial pyramid of dilated convolutions (SPC).}
	\label{tab:ablationStudies}
\end{table}	

\noindent \textbf{ReLU vs PReLU:} (Table \ref{tab:actiImp}) Replacing ReLU \cite{nair2010rectified} with PReLU \cite{he2015delving} in ESPNet-A improved the accuracy by $2\%$, while having a minimal impact on the network complexity.

\noindent \textbf{Residual learning in ESP:} (Table \ref{tab:blockImp}) The accuracy of ESPNet-A dropped by about 2\% when skip-connections in ESP (\Fig \ref{fig:sppBlocks}) modules were removed. This verifies the effectiveness of the residual learning.

\noindent \textbf{Down-sampling:} (Table \ref{tab:downImp}) Replacing the standard strided convolution with the strided ESP in ESPNet-A improved accuracy by 1\% with $33\%$ parameter reduction.

\noindent \textbf{Width divider ($K$):} (Table \ref{tab:parallelImp}) Increasing $K$ enlarges the  effective receptive field of the ESP module, while simultaneously decreasing the number of network parameters. Importantly, ESPNet-A's accuracy decreased with increasing $K$. For example, raising $K$ from 2 to 8 caused  ESPNet-A's accuracy to drop by 11\%. This drop in accuracy is explained in part by the ESP module's effective receptive field growing beyond the size of its input feature maps. For an image with size $1024 \times 512$, the spatial dimensions of the input feature maps at spatial level $l=2$ and $l=3$ are $256 \times 128$ and $128 \times 64$, respectively. However, some of the kernels have larger receptive fields ($257 \times 257$ for $K=8$). The weights of such kernels do not contribute to learning, thus resulting in lower accuracy. At $K=5$, we found a good trade-off between number of parameters and accuracy, and therefore, we used $K=5$ in our experiments.

\noindent \textbf{ESPNet-A $\rightarrow$ ESPNet-C:} (Table \ref{tab:modelImp}) Replacing the convolution-based network width expansion operation in ESPNet-A with the concatenation operation in ESPNet-B improved the accuracy by about $1\%$ and did not increase the number of network parameters noticeably. With input reinforcement (ESPNet-C), the accuracy of ESPNet-B further improved by about 2\%, while not increasing the network parameters drastically. This is likely due to the fact that the input reinforcement method establishes a direct link between the input image and encoding stage, improving the flow of information. 

The closest work to our input reinforcement method is the input-aware fusion method of \cite{mehta2017learning}, which learns representations on the down-sampled input image and additively combines them with the convolutional unit. When  the proposed input reinforcement method was replaced with the input-aware fusion in \cite{mehta2017learning}, no improvement in accuracy was observed, but the number of network parameters increased by about 10\%. 

\noindent \textbf{ESPNet-C vs ESPNet:} (Table \ref{tab:espnetCompare}) Adding a light-weight decoder to ESPNet-C improved the accuracy by about 6\%, while increasing the number of parameters and network size by merely 20,000 and 0.06 MB from ESPNet-C to ESPNet, respectively.

\noindent \textbf{Impact of different convolutions in the ESP block:} The ESP block uses point-wise convolutions for reducing the high-dimensional feature maps to low-dimensional space and then transforms those feature maps using a spatial pyramid of dilated convolutions (SPCs) (see Sec. 3). To understand the influence of these two components, we performed the following experiments. 
\textit{1) Point-wise convolutions:} We replaced point-wise convolutions with $3\times 3$ standard convolutions in the ESP block (see C1 and C2 in Table \ref{tab:effect}), and the resultant network demanded more resources (e.g., 47\% more parameters) while improving the mIOU by 1.8\%, showing that point-wise convolutions are effective. Moreover, the decrease in number of parameters due to point-wise convolutions in the ESP block enables the construction of deep and efficient networks (see Table \ref{tab:espnetCompare}). 
\textit{2) SPCs:} We replaced $3\times 3$ dilated convolutions with $3\times 3$ standard convolutions in the ESP block. Though the resultant network is as efficient as with dilated convolutions, it is 1.6\% less accurate; suggesting SPCs are effective (see C2 and C3 in Table \ref{tab:effect}).

\section{Conclusion}
We introduced a semantic segmentation network, ESPNet, based on an efficient spatial pyramid module. In addition to legacy metrics, we introduced several new system-level metrics that help to analyze the performance of a CNN network. Our empirical analysis suggests that ESPNets are fast and efficient. We also demonstrated that ESPNet learns good generalizable representations of the objects and perform well in the wild. 

\begin{small}
	\vspace{0.2cm}
\noindent {\bf Acknowledgement:} This research was supported by the Intelligence Advanced Research Projects Activity (IARPA) via Interior/Interior Business Center (DOI/IBC) contract number D17PC00343, the Washington  State  Department  of  Transportation  research grant T1461-47, NSF III (1703166), the National Cancer Institute awards (R01 CA172343, R01 CA140560, and RO1 CA200690), Allen Distinguished Investigator Award, Samsung GRO award, and gifts from Google, Amazon, and Bloomberg. We would also like to acknowledge NVIDIA Corporation for donating the Jetson TX2 board and the Titan X Pascal GPU used for this research. We also thank the anonymous reviewers for their helpful comments. The U.S. Government is authorized to reproduce and distribute reprints for Governmental purposes notwithstanding any copyright annotation thereon. Disclaimer: The views and conclusions contained herein are those of the authors and should not be interpreted as necessarily representing endorsements, either expressed or implied, of IARPA, DOI/IBC, or the U.S. Government.
\end{small}

\bibliographystyle{splncs}
\bibliography{main}

\clearpage

\appendix

\section{Hardware Details}
\label{sec:hardwareDetailsSup}
Three machines were used in our experiments. Table \ref{tab:hardDetailsSup} summarizes the details about these machines. A computing platform (e.g. Jetson TX2) on an edge device shares the global memory or RAM between CPU and GPU, while laptop and desktop devices have dedicated CPU and GPU memory.

NVIDIA Jetson TX2 can run in different modes. In performance mode (Max-P), all CPU cores are enabled in TX2, while in normal mode (Max-Q mode) only 4 out of 6 CPU cores are active. CPU and GPU clock frequencies are different in these modes and therefore, applications will have different power requirements in different modes.

\begin{table}[b!]
\centering
\resizebox{\columnwidth}{!}{
\begin{tabular}{c|l|c|c|c}
\toprule
& & \textbf{Desktop} & \textbf{Laptop} & \textbf{Edge Device}\\
\midrule
& \textbf{CPU Architecture} & x86\_64& x86\_64  & aarch64 \\
 \cline{2-5}
\multirow{6}{*}{\textbf{CPU}} & \multirow{2}{*}{\textbf{CPU Cores}} & \multirow{2}{*}{8} & \multirow{2}{*}{8} & 6 (Max-P mode) \\ 
 & &  &  & 4 (Max-Q mode) \\ 
 \cline{2-5}
& \multirow{2}{*}{\textbf{CPU Model Name}} & Intel(R) Core(TM) & Intel(R) Core(TM)  & ARMv8 Processor  \\
&  &  i7-6700k @ 4 GHz &  i7-6700HQ CPU @ 2.60 GHz &  rev 3(v81) \\
 \cline{2-5}
& \textbf{L1 Cache} & 32 KB & 32 KB & 32 KB\\
 \cline{2-5}
& \textbf{L2 Cache} & 256 KB & 256 KB & 2 MB \\
 \cline{2-5}
& \textbf{L3 Cache} & 8 MB & 6 MB & -- \\
 \cline{2-5}
& \textbf{RAM} & 16 GB & 16 GB & 8 GB (shared) \\
\midrule
\multirow{9}{*}{\textbf{GPU}} & \textbf{GPU Model Name} & TitanX Pascal &  GeForce GTX 960M & Tegra X2\\
 \cline{2-5}
& \textbf{CUDA Driver Version} & 9.1 & 9.1 & 8.0 \\
 \cline{2-5}
& \textbf{Global Memory} & 12 GB  & 4 GB  & 8 GB (shared) \\
 \cline{2-5}
& \multirow{2}{*}{\textbf{Max. GPU frequency}} & 1.53 GHz & 1.18 GHz & 1.3 GHz (Max-P mode))\\
&  &  & &824 MHz (Max-Q mode)\\
 \cline{2-5}
& \textbf{CUDA Cores} & 3584 &  640 & 256\\
 \cline{2-5}
& \textbf{Streaming multiprocessors (SM)} & 28 & 5 & 2\\
 \cline{2-5}
& \textbf{CUDA Cores per SM} & 128 & 128 & 128 \\
\bottomrule
\end{tabular}
}
\caption{This table summarizes the hardware that we used in our experiments.}
\label{tab:hardDetailsSup}
\end{table}

\section{The path from ESPNet-A to ESPNet}
\label{sec:detailsBlocks}
\begin{figure}[t!]
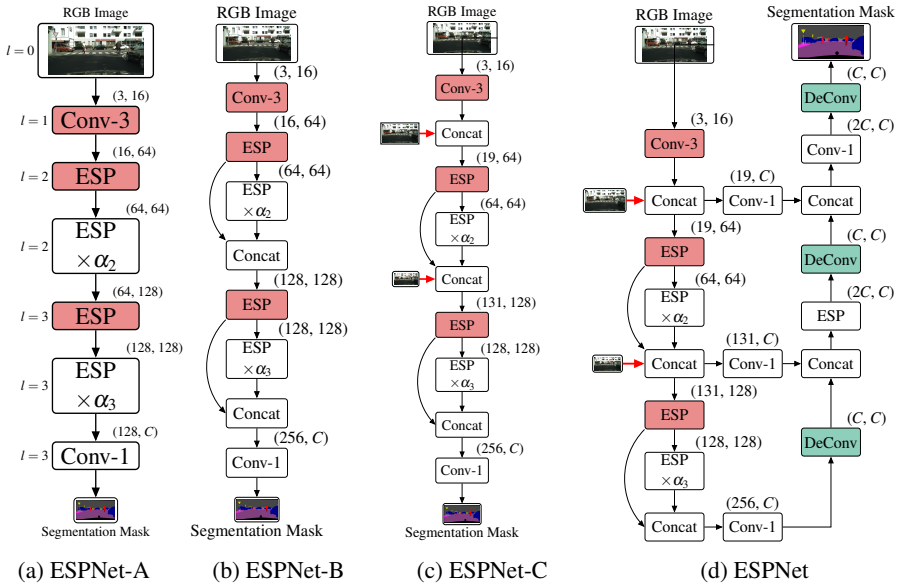

\centering
\begin{subfigure}[b]{0.18\columnwidth}
\resizebox{!}{200px}{
\hspace*{-70pt}\input{Digrams.tikz}\architectureEncA
}
\caption{ESPNet-A}
\label{fig:archASup}
\end{subfigure}
\hfill
\begin{subfigure}[b]{0.2\columnwidth}
\resizebox{!}{200px}{
\hspace*{-70pt}\input{Digrams.tikz}\architectureEncB
}
\caption{ESPNet-B}
\label{fig:archBSup}
\end{subfigure}
\hfill
\begin{subfigure}[b]{0.2\columnwidth}
\resizebox{!}{200px}{
\hspace*{-90pt}\input{Digrams.tikz}\architectureEncC
}
\caption{ESPNet-C}
\label{fig:archCSup}
\end{subfigure}
\hfill
\begin{subfigure}[b]{0.35\columnwidth}
\resizebox{!}{200px}{
\hspace*{-90pt}\input{Digrams.tikz}\architecture
}
\caption{ESPNet}
\label{fig:espnetFullArchSup}
\end{subfigure}
\caption{The path from ESPNet-A to ESPNet. Red and green color boxes represent the modules responsible for down-sampling and up-sampling operations, respectively. Spatial-level $l$ is indicated on the left of every module in (a). We denote each module as (\# input channels, \# output channels). Here, Conv-$n$ represents $n\times n$ convolution. This figure is the same as \Fig \ref{fig:espNetVariants}.}
\label{fig:espNetVariantsSup}
\end{figure}

Different variants of ESPNet are shown in \Fig \ref{fig:espNetVariantsSup}. The first variant, ESPNet-A (\Fig \ref{fig:archASup}), is a standard network that takes an RGB image as an input and learns representations at different spatial levels using the ESP module to produce a segmentation mask. The second variant, ESPNet-B (\Fig \ref{fig:archBSup}), improves the flow of information inside ESPNet-A by sharing the feature maps between the previous strided ESP module and the previous ESP module. The third variant, ESPNet-C (\Fig \ref{fig:archCSup}), reinforces the input image inside ESPNet-B to further improve the flow of information. These three variants produce outputs whose spatial dimensions are $\frac{1}{8}{th}$ of the input image. The fourth variant, ESPNet (\Fig \ref{fig:espnetFullArchSup}), adds a light weight decoder (built using a principle of \textit{reduce-upsample-merge}) to ESPNet-C that outputs the segmentation mask of the same spatial resolution as the input image. The building block functions used to build the ESPNet (from ESPNet-A to ESPNet) are discussed next.

\noindent \textbf{Efficient down-sampling:} Recent CNN architectures have used strided convolution (e.g. \cite{springenberg2014striving,he2016deep,xie2017aggregated}) instead of pooling operations (e.g. \cite{simonyan2014very,krizhevsky2012imagenet}) for down-sampling operations, because it allows the non-linear down-sampling operations to be learned while simultaneously enabling expansion of the network width. Standard strided convolutional operations are expensive; therefore, they are replaced by strided ESP modules for down-sampling. Point-wise convolutions are replaced by  $n\times n$ strided convolutions in the ESP module for learning non-linear down-sampling operations. The spatial dimensions of the feature maps are changed by down-sampling operations. Following \cite{he2016deep,huang2017densely}, we do not combine the input and output feature maps using the skip-connection during down-sampling operations. The number of parameters learned by strided convolution and strided ESP are $n^2MN$ and $\frac{n^2MN}{K} + \left(\frac{n^2N^2}{K^2} \cdot K\right)$, respectively. By expressing strided convolution as strided ESP for down-sampling, the number of parameters required is reduced by a factor of $\frac{KM}{M + N}$ and the effective receptive field is increased by $\sim [2^{K-1}]^2$ times. We will refer to this network as ESPNet-A (\Fig \ref{fig:archASup}).

\noindent \textbf{Network width expansion:} To maintain the computational complexity at each spatial level, traditional CNNs (e.g. \cite{simonyan2014very,he2016deep,xie2017aggregated}) double the width of the network after every down-sampling operation, usually using a convolution operation. Following \cite{huang2017densely}, we concatenate the feature maps received from the previous strided ESP module and the previous ESP module to increase the width of the network, as shown in \Fig \ref{fig:archBSup} with a curved arrow. The concatenation operation establishes a long-range connection between the input and output at the same spatial level and, therefore, improves the flow of information inside the network. We will refer to this network as ESPNet-B (\Fig \ref{fig:archBSup}). 

\noindent \textbf{Input reinforcement:} Spatial information is lost due to down-sampling and convolutional operations. To compensate, we reinforce the input image inside the network. We down-sample the input-image and concatenate it with the feature maps from the previous strided ESP module and the previous ESP module. We will refer to ESPNet-B with input reinforcement as ESPNet-C (\Fig \ref{fig:archCSup}). Since the input RGB image has only 3 channels, the increase in network complexity due to input reinforcement is minimal.

\noindent \textbf{Depth multiplier $\alpha$:} To build deeper computationally efficient networks for edge devices without changing the network topology, we introduce a hyper-parameter $\alpha$ to control the depth of the network. This parameter, $\alpha$, repeats the ESP module $\alpha_l$ times at spatial level $l$. CNNs require more memory at higher spatial levels i.e. at $l=0$ and $l=1$ because of the high spatial dimensions of feature maps at these levels. To be memory efficient, we do not repeat ESP or convolutional modules at these spatial levels.

\begin{figure}[t!]
\centering
\includegraphics[width=0.6\textwidth]{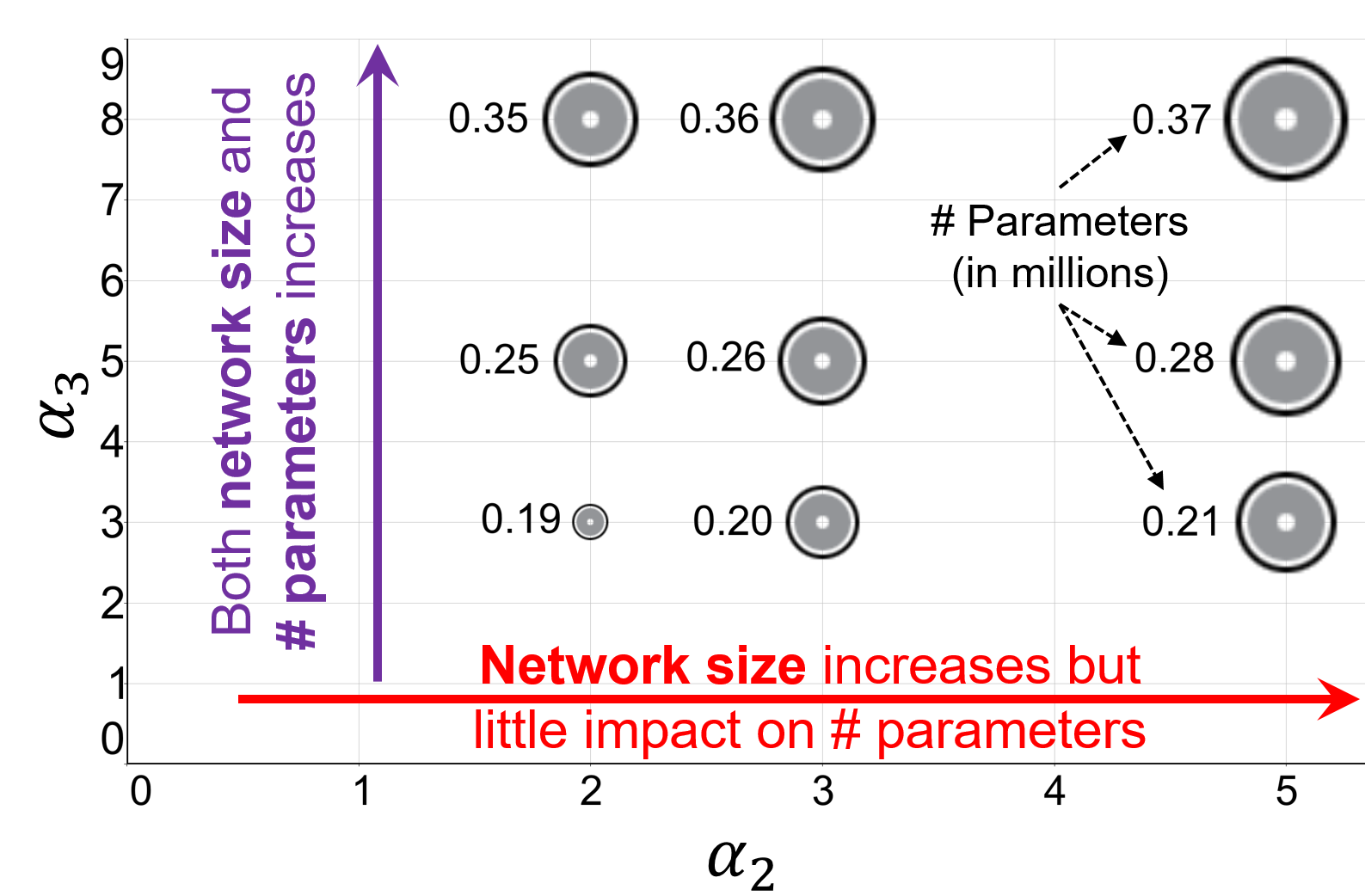}
\caption{Relationship between depth multipliers $\alpha_2$ and $\alpha_3$ for creating efficient networks. Here, circle size $\propto$ network size.}
\label{fig:resourceReqSup}
\end{figure}

As we change the values of these parameters, the amount of computational resources required by a network will change. \Fig \ref{fig:resourceReqSup} shows the impact of $\alpha_l, l=\{2, 3\}$  on the network parameters and its size. As we increase $\alpha_2$, the network size increases with little impact on the number of parameters. When we increase $\alpha_3$, both the network size and number of parameters increase. Both the number of parameters and network size should increase with depth \cite{he2016deep,xie2017aggregated,SzegedyIV16InceptionV4}. Therefore, for creating deep and efficient ESPNet networks, we fix the value of $\alpha_2$ and vary the value of $\alpha_3$.

\noindent \textbf{RUM for efficient decoding:} The spatial resolution of the output produced by ESPNet-C is $\frac{1}{8}{th}$ of the input image size. Up-sampling the feature maps directly, say using bilinear interpolation, may give good accuracy  on a standard metric, but the output is usually coarse \cite{long2015fully}. We adopt a bottom-up approach (e.g. \cite{badrinarayanan2017segnet,ronneberger2015u}) to aggregate the multi-level information learned by ESPNet-C using a simple rule: \textit{Reduce-Upsample-Merge} (RUM). \textit{Reduce:} The feature map from spatial levels $l$ and $l-1$ are projected to a $C$-dimensional space, where $C$ represents the number of classes in the dataset. \textit{Upsample:} The reduced feature map from spatial level $l$ is upsampled by a factor of $2$ using a $2\times2$ deconvolutional kernel so that it has the same spatial dimensions as that of the feature map at level $l-1$. \textit{Merge:} The up-sampled feature map from level $l$ is then combined with the $C$-dimensional feature map from level $l-1$ using a concatenation operation. This process is repeated until the spatial dimensions of the feature map are the same as the input image. We refer to this network as ESPNet (\Fig \ref{fig:espnetFullArchSup}).

\section{Image Size vs. Inference Speed}
\label{sec:impactSup}
\begin{figure}[b!]
	\centering
	\includegraphics[width=0.65\columnwidth]{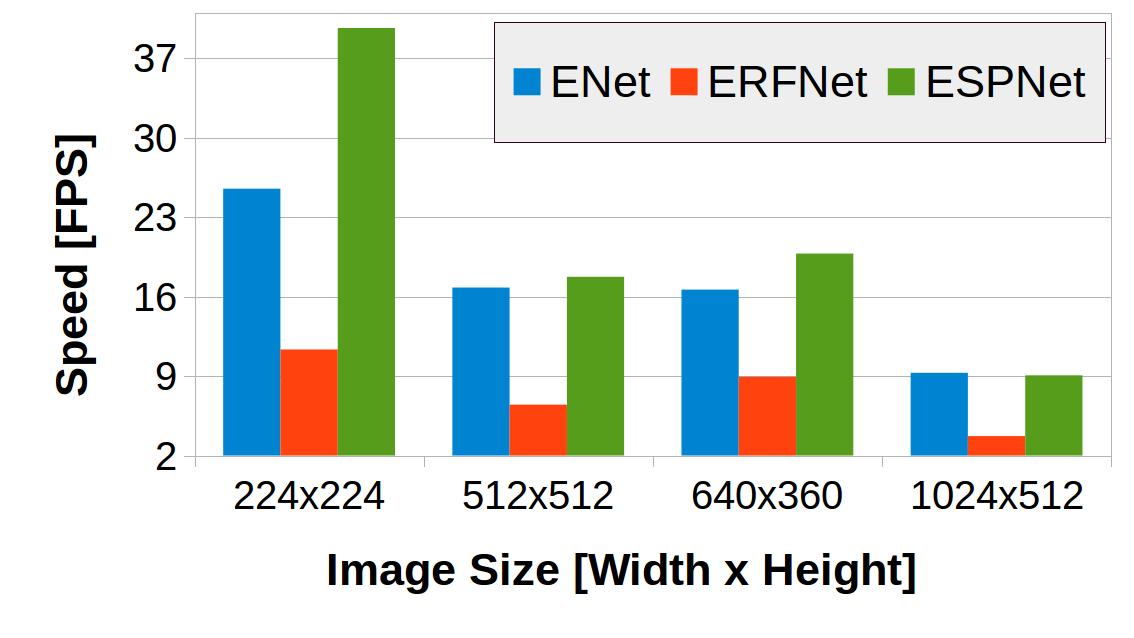}
	\caption{The impact of image size on the inference speed on an edge device}
	\label{fig:impactImSpeedSup}
\end{figure}

Figure \ref{fig:impactImSpeedSup} summarizes the impact of image size on the inference speed. At smaller image resolutions (224x224 and 640x360), ESPNet is faster than ENet and ERFNet. However, ESPNet delivers a similar inference speed to ENet for-high resolution images. We presume that ESPNet is bottlenecked by the limited and shared resources on the TX2 device. We note that ESPNet processes high resolution images faster than ENet on high-end devices, such as laptop and desktop.

\section{Top-10 Kernels in ESPNet, ENet, and ERFNet}
\label{sec:topKernSUp}

\begin{figure}[t!]
	\centering
	\begin{subfigure}[b]{0.49\columnwidth}
		\centering
		\includegraphics[width=\columnwidth]{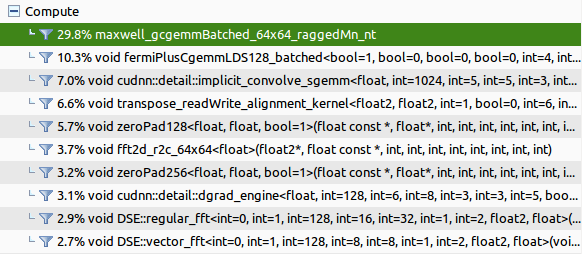}
		\caption{ENet}
		\label{fig:enetKernSup}
	\end{subfigure}
	\hfill
	\begin{subfigure}[b]{0.49\columnwidth}
		\centering
		\includegraphics[width=\columnwidth]{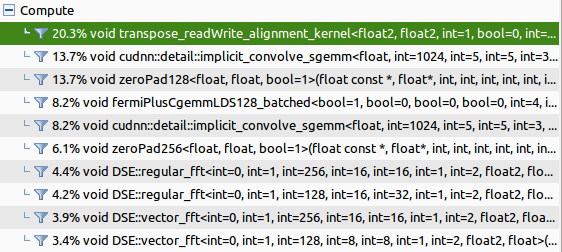}
		\caption{ERFNet}
		\label{fig:erfKernSup}
	\end{subfigure}
	\vfill
	\begin{subfigure}[b]{0.49\columnwidth}
		\centering
		\includegraphics[width=\columnwidth]{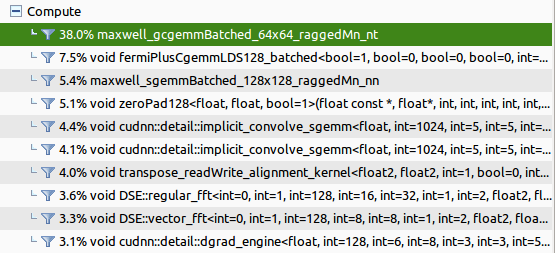}
		\caption{ESPNet ($\alpha_2=2, \alpha_3=3$)}
		\label{fig:espOneSup}
	\end{subfigure}
	\hfill
	\begin{subfigure}[b]{0.49\columnwidth}
		\centering
		\includegraphics[width=\columnwidth]{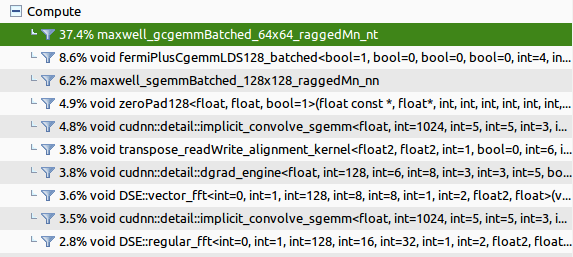}
		\caption{ESPNet ($\alpha_2=2, \alpha_3=5$)}
		\label{fig:espTwoSup}
	\end{subfigure}
	\vfill
	\begin{subfigure}[b]{\columnwidth}
		\centering
		\includegraphics[width=0.49\columnwidth]{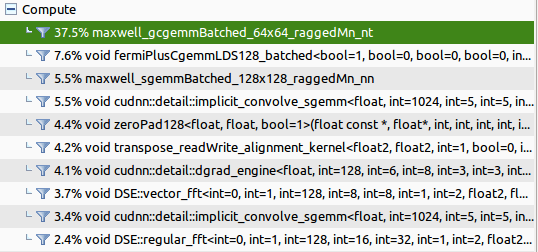}
		\caption{ESPNet ($\alpha_2=2, \alpha_3=8$)}
		\label{fig:espThreeSup}
	\end{subfigure}
	\caption{This figure visualizes the top-10 kernels along with their contribution towards compute resource utilization. The top-1 kernel is highlighted in green color.}
	\label{fig:topKernelVisSup}
\end{figure}

Convolutional operations are implemented using a highly optimized general matrix multiplication (GEMM) operations and memory re-ordering operations such as im2col. For fast and efficient networks, the kernel corresponding to GEMM operations should have high contribution towards compute resource utilization. Figure \ref{fig:topKernelVisSup} visualizes the top-10 kernels executed by ENet, ERFNet, and ESPNet. We can see that the top-1 kernel in ESPNet is GEMM, and it is responsible for about 38\% of the total computational time. Since convolution operations are implemented using the GEMM kernel, this suggest that ESPNet utilizes the limited computational resources available in TX2 efficiently. Similarly, the top-1 kernel in ENet is also GEMM; however, the contribution of this kernel towards computing is not as high as ESPNet. This is why the sensitivity of ENet towards GPU frequency is low and runs $1.27 \times$ slower on NVIDIA TitanX than ESPNet while running at almost the same rate on the NVIDIA TX2. On the other hand, the top-1 kernel in ERFNet is the memory alignment kernel. This suggests that ERFNet gets bottlenecked by the memory operations.

\section{Resource Utilization Plots for ENet, ERFNet, and ESPNet}
\label{sec:resourceUtilTop}

\begin{figure}[t!]
\centering
\includegraphics[width=\columnwidth]{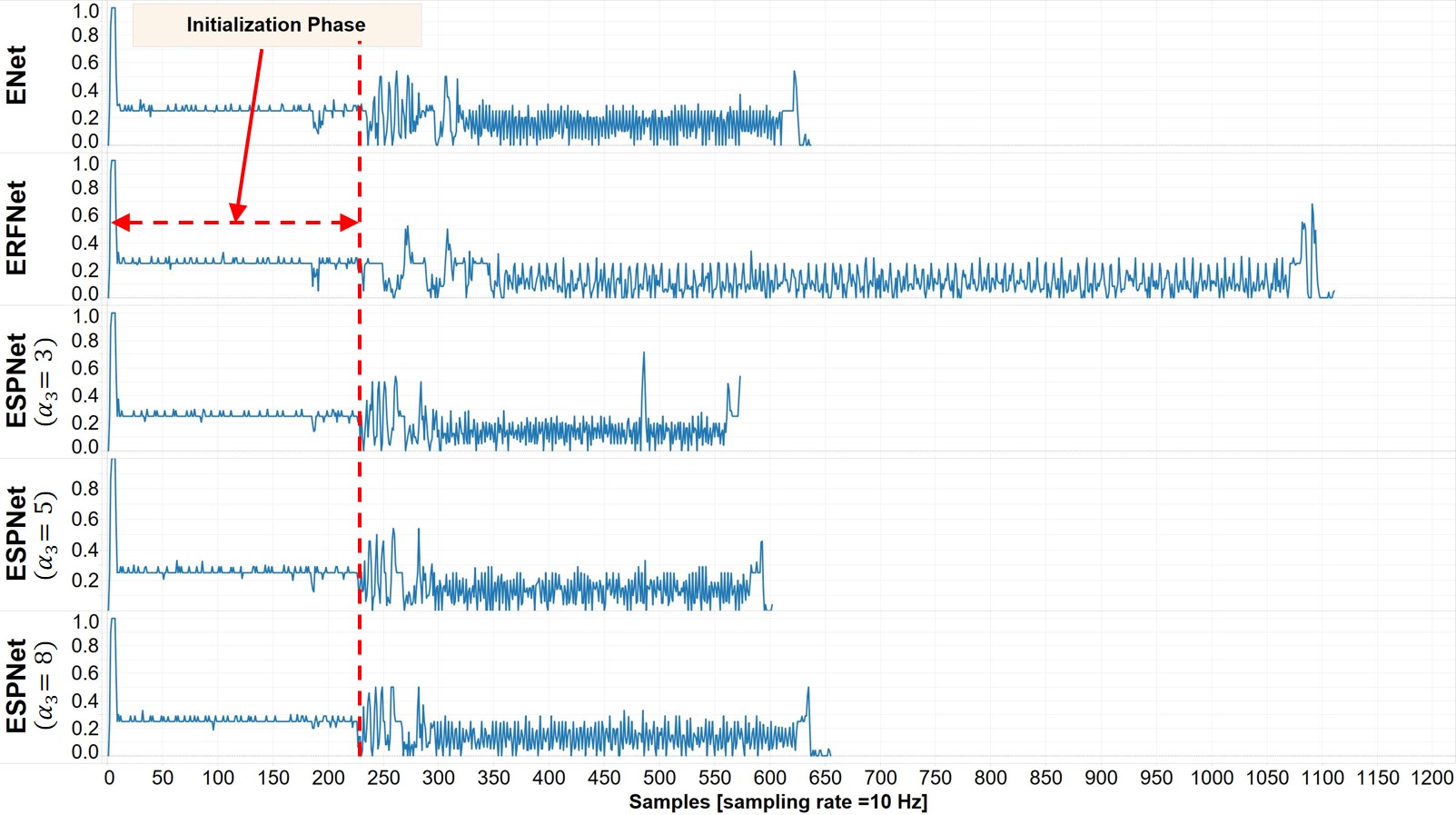}
\caption{This figure compares the \textbf{CPU utilization} rates on NVIDIA Jetson TX2. For ESPNet, we used $\alpha_2 = 2$. Here, 1.0 represents 100\% CPU utilization.}
\label{fig:cpuUtilSupp}
\end{figure}

\begin{figure}[t!]
\centering
\includegraphics[width=\columnwidth]{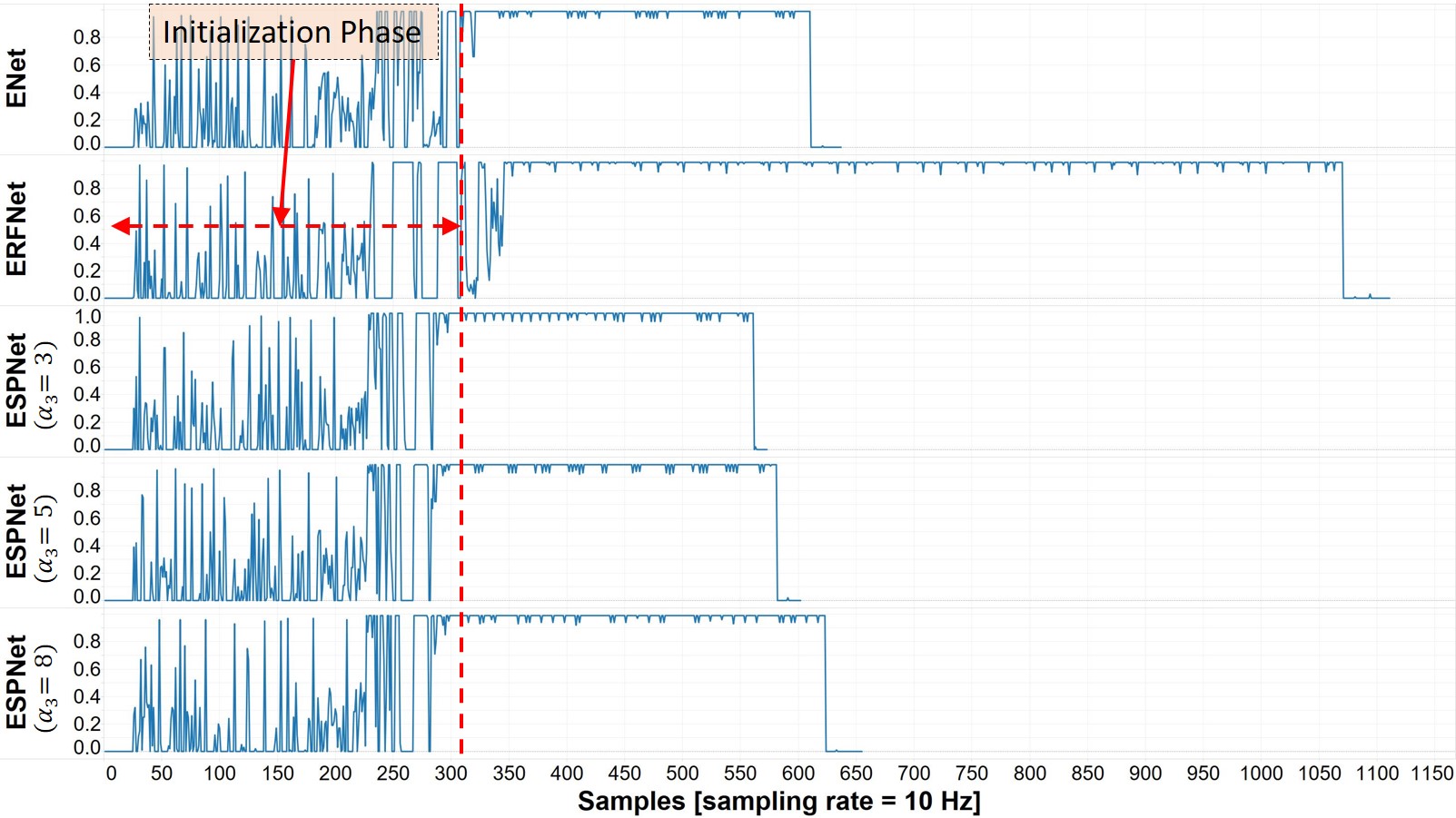}
\caption{This figure compares the \textbf{GPU utilization} rates on NVIDIA Jetson TX2. For ESPNet, we used $\alpha_2 = 2$. Here, 1.0 represents 100\% GPU utilization. }
\label{fig:gpuUtilSupp}
\end{figure}

\begin{figure}[t!]
\centering
\includegraphics[width=0.75\columnwidth]{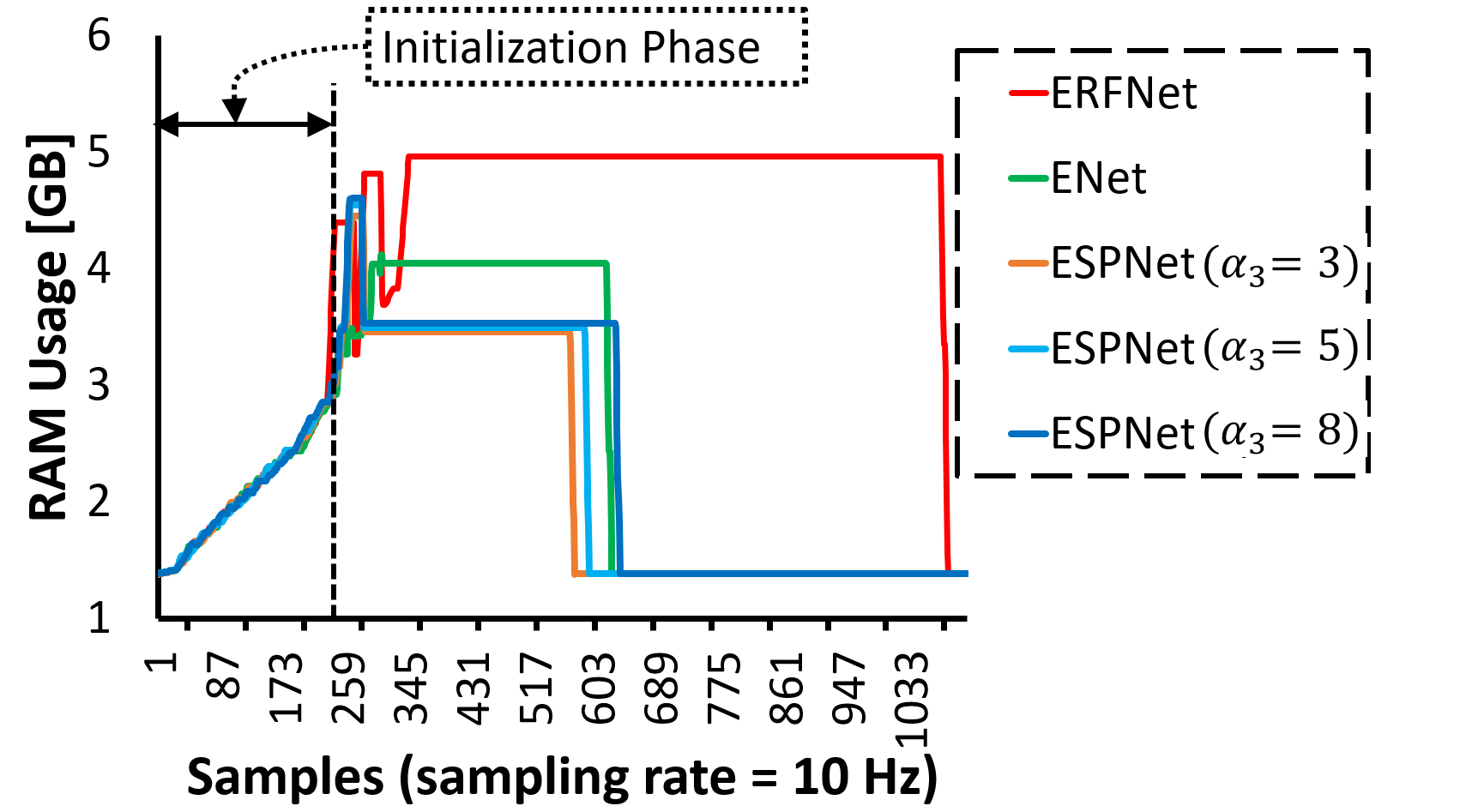}
\caption{This figure compares the \textbf{memory utilization} on NVIDIA Jetson TX2. For ESPNet, we used $\alpha_2 = 2$. Maximum available memory on TX2 is 8 GB and is \textit{shared} between CPU and GPU.}
\label{fig:memUtilSupp}
\end{figure}

Figures \ref{fig:cpuUtilSupp},  \ref{fig:gpuUtilSupp}, and  \ref{fig:memUtilSupp} show the utilization of TX2 resources (CPU, GPU, and memory) over time for ENet, ERFNet and ESPNet. The data were collected using Tegrastats in \textit{Max-Q} mode. These networks are throughput intensive, and therefore, GPU utilization rates are high while CPU utilization rates are low for these networks. Note that the average CPU utilization rate is below 25\%; suggesting that these networks are using only one CPU core out of the available four CPU cores and can be bound to a single CPU core for better utilization of CPU resources, if running additional applications on TX2. Memory utilization rates are significantly different for these networks. The memory footprint of ESPNet is low in comparison to ENet and ERFNet, suggesting ESPNet is suitable for memory constrained devices.

Recall that ESPNet with $\alpha_2 = 2$ and $\alpha_3=8$ learns the same number of parameters as ENet. However, ESPNet has a low memory footprint than ENet (\Fig \ref{fig:memUtilSupp}); suggesting ESPNet is more memory efficient and utilizes the shared memory efficiently.

\clearpage 

\section{Results on the Cityscape and the Mapillary Dataset}
\label{sec:resultsSup}
\begin{figure}[t!]
\begin{subfigure}[b]{\columnwidth}
\includegraphics[width=\columnwidth]{images/legend.jpg}
\end{subfigure}
\begin{subfigure}[b]{\columnwidth}
\includegraphics[width=\columnwidth]{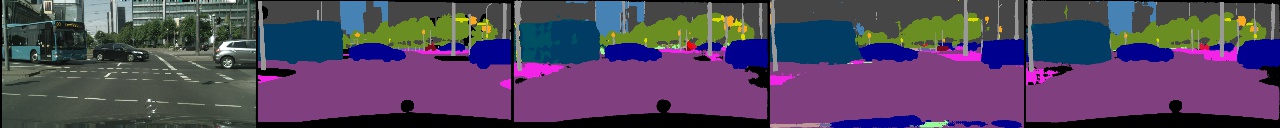}
\end{subfigure}
\begin{subfigure}[b]{\columnwidth}
\includegraphics[width=\columnwidth]{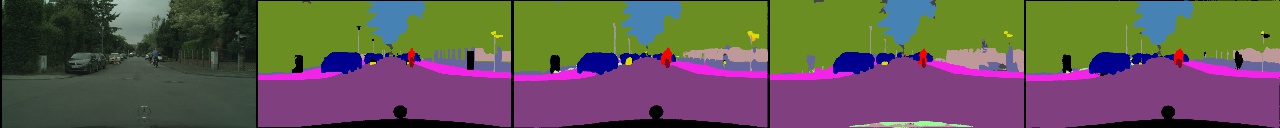}
\end{subfigure}
\begin{subfigure}[b]{\columnwidth}
\includegraphics[width=\columnwidth]{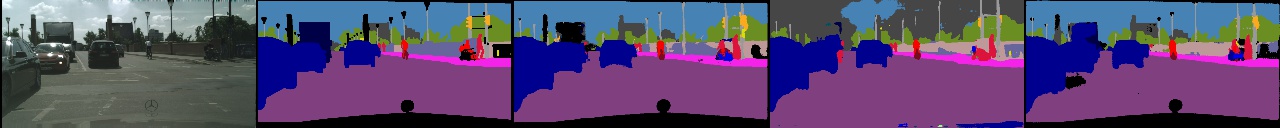}
\end{subfigure}
\begin{subfigure}[b]{\columnwidth}
\includegraphics[width=\columnwidth]{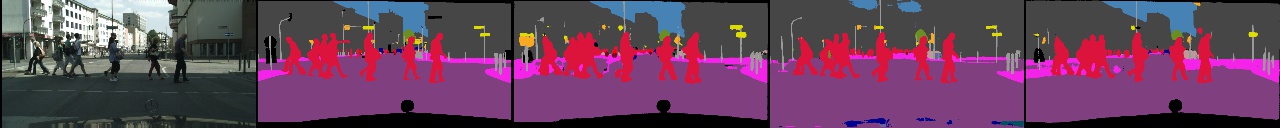}
\end{subfigure}
\begin{subfigure}[b]{\columnwidth}
\includegraphics[width=\columnwidth]{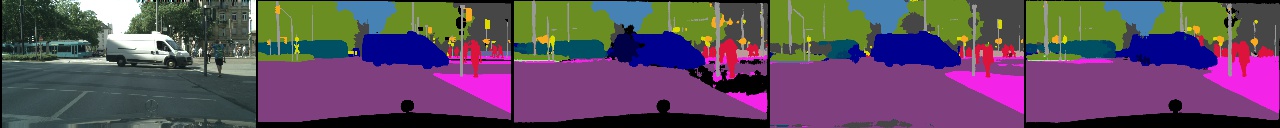}
\end{subfigure}
\begin{subfigure}[b]{\columnwidth}
\includegraphics[width=\columnwidth]{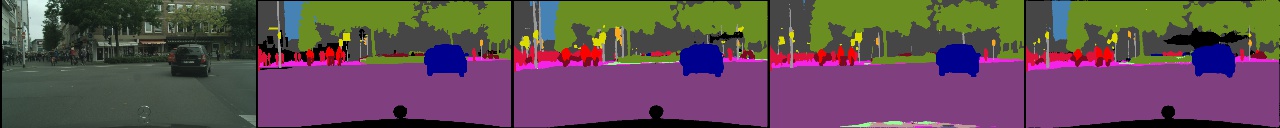}
\end{subfigure}
\begin{subfigure}[b]{\columnwidth}
\includegraphics[width=\columnwidth]{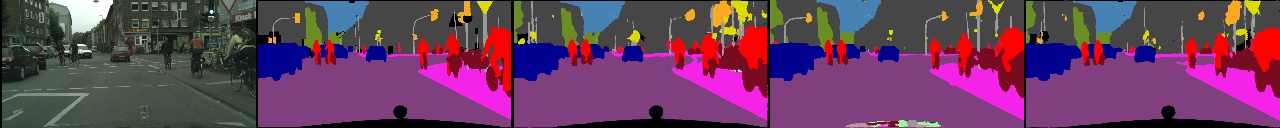}
\end{subfigure}
\begin{subfigure}[b]{\columnwidth}
\includegraphics[width=\columnwidth]{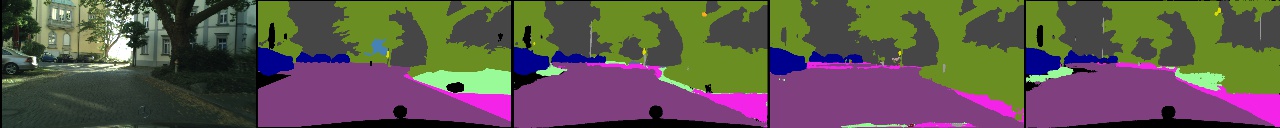}
\end{subfigure}
\begin{subfigure}[b]{\columnwidth}
\includegraphics[width=\columnwidth]{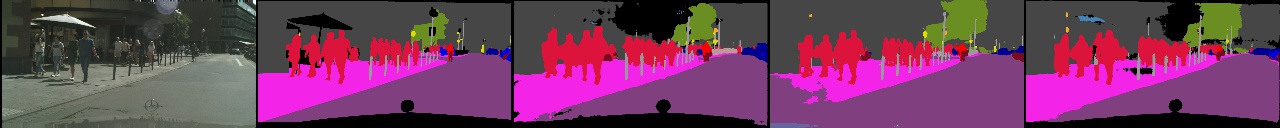}
\end{subfigure}
\begin{subfigure}[b]{\columnwidth}
\includegraphics[width=\columnwidth]{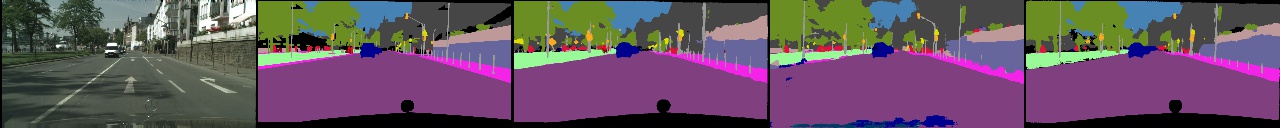}
\end{subfigure}
\begin{subfigure}[b]{\columnwidth}
\includegraphics[width=\columnwidth]{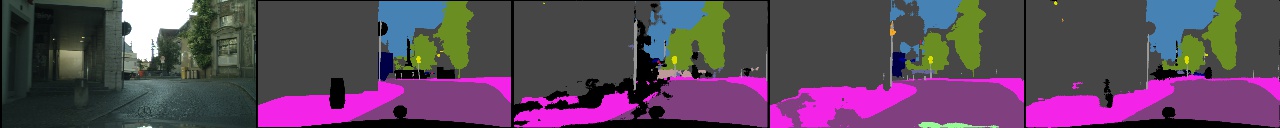}
\end{subfigure}
\begin{subfigure}[b]{\columnwidth}
\includegraphics[width=\columnwidth]{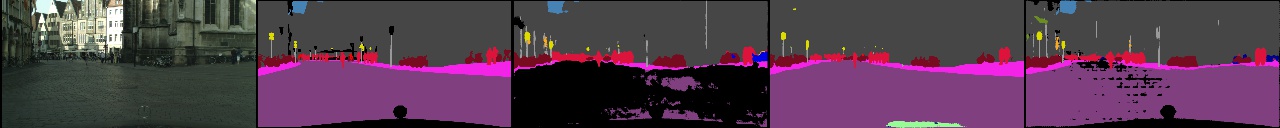}
\end{subfigure}
\caption{Qualitative results on the Cityscape validation dataset.}
\label{fig:cityResultsSupp}
\end{figure}

\begin{figure}[h!]
\begin{subfigure}[b]{\columnwidth}
\includegraphics[width=\columnwidth]{images/legend.jpg}
\end{subfigure}
\begin{subfigure}[b]{\columnwidth}
\includegraphics[width=\columnwidth]{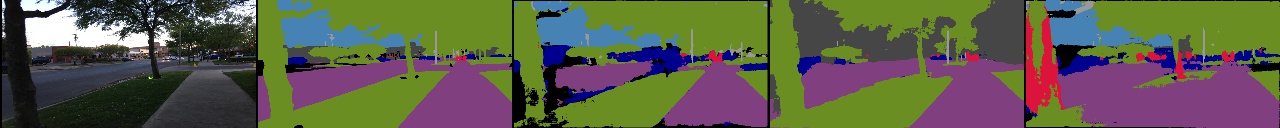}
\end{subfigure}
\begin{subfigure}[b]{\columnwidth}
\includegraphics[width=\columnwidth]{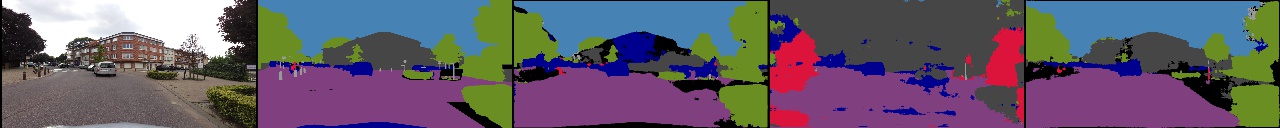}
\end{subfigure}
\begin{subfigure}[b]{\columnwidth}
\includegraphics[width=\columnwidth]{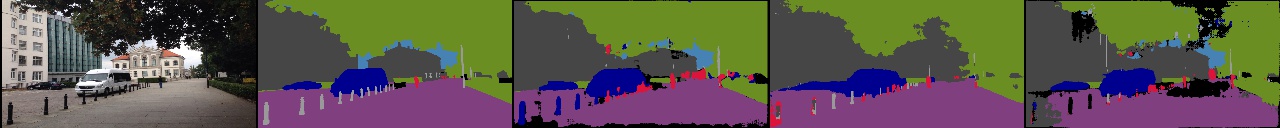}
\end{subfigure}
\begin{subfigure}[b]{\columnwidth}
\includegraphics[width=\columnwidth]{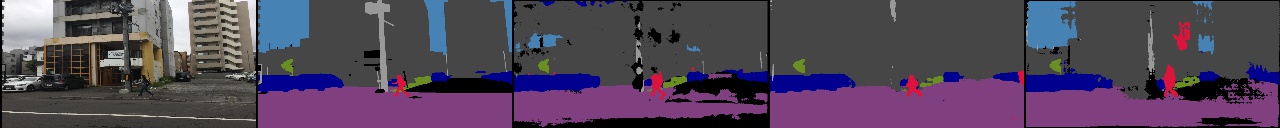}
\end{subfigure}
\begin{subfigure}[b]{\columnwidth}
\includegraphics[width=\columnwidth]{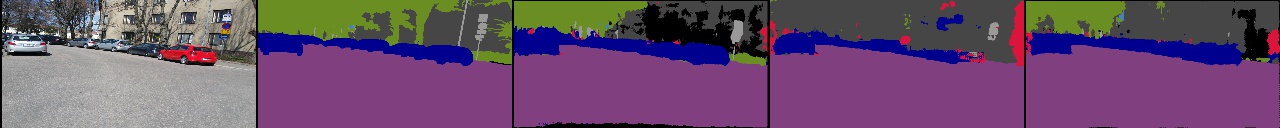}
\end{subfigure}
\begin{subfigure}[b]{\columnwidth}
\includegraphics[width=\columnwidth]{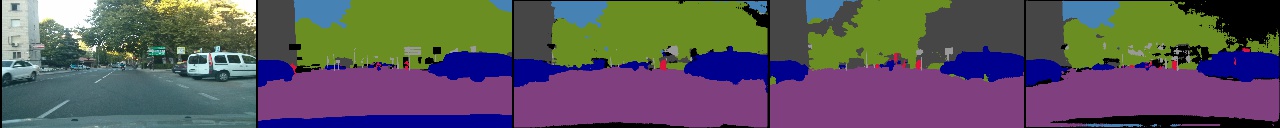}
\end{subfigure}
\begin{subfigure}[b]{\columnwidth}
\includegraphics[width=\columnwidth]{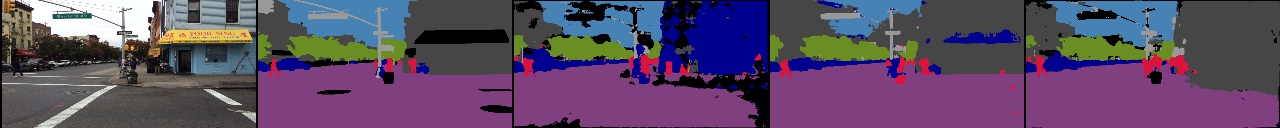}
\end{subfigure}
\begin{subfigure}[b]{\columnwidth}
\includegraphics[width=\columnwidth]{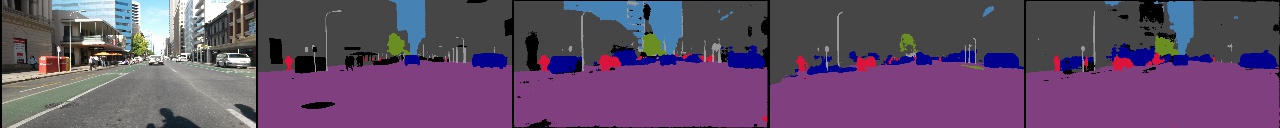}
\end{subfigure}
\begin{subfigure}[b]{\columnwidth}
\includegraphics[width=\columnwidth]{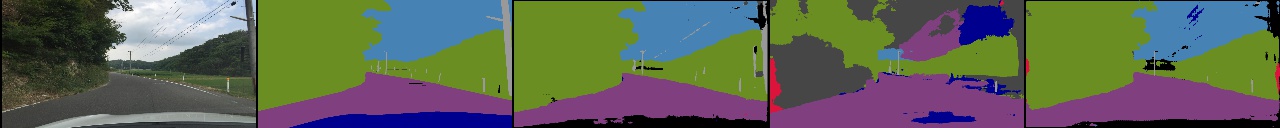}
\end{subfigure}
\begin{subfigure}[b]{\columnwidth}
\includegraphics[width=\columnwidth]{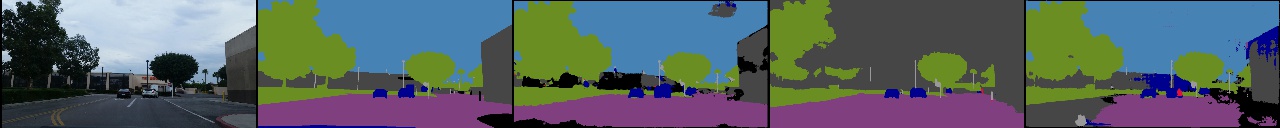}
\end{subfigure}
\begin{subfigure}[b]{\columnwidth}
\includegraphics[width=\columnwidth]{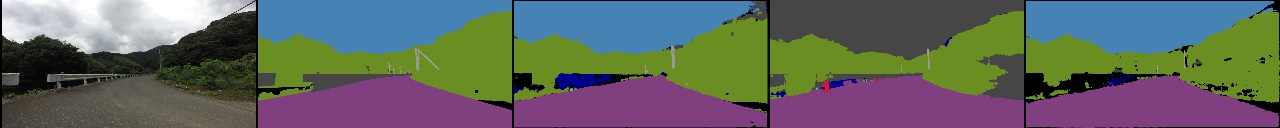}
\end{subfigure}
\begin{subfigure}[b]{\columnwidth}
\includegraphics[width=\columnwidth]{mapillaryResults/sample_4.jpg}
\end{subfigure}
\caption{Qualitative results on the Mapillary validation dataset.}
\label{fig:mapResultsSupp}
\end{figure}

A summary of class-wise and category-wise results on the Cityscape \cite{cordts2016cityscapes} dataset was given in Table \ref{tab:compareClassCatSup}, while category-wise results on the Mapillary \cite{MVD2017} dataset were given in Table \ref{tab:compareCatMapSup}. Though ERFNet outperformed ENet and ESPNet on every class, it performed badly on the Mapillary dataset. In particular, ERFNet struggled classifying simple classes, such as sky, on the Mapillary dataset, while on such classes, ENet and ESPNet performed relatively well. We note that ESPNet learns good generalization representations about the objects and performs well, even in the wild. Qualitative results on the Cityscape and Mapillary dataset were given in Figure \ref{fig:cityResultsSupp} and Figure \ref{fig:mapResultsSupp}, respectively. 

\clearpage

\begin{figure}[t!]
\centering
\includegraphics[width=0.7\columnwidth]{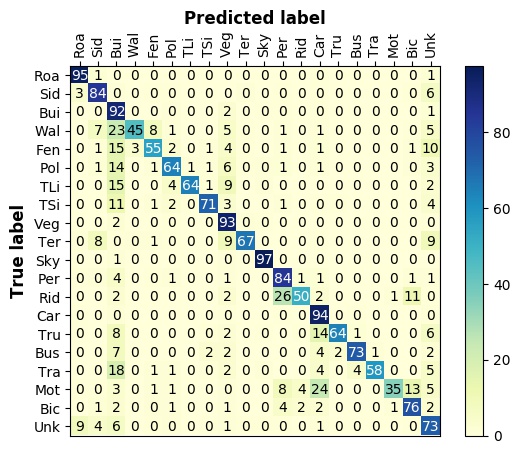}
\caption{ESPNet's (with $\alpha_2 = 2$ and $\alpha_3 = 8$) confusion matrix on the Cityscape \textit{validation} set. ESPNet makes some mistakes between classes that belong to the same category, and hence has lower class-wise accuracy. However, ESPNet delivers a good category-wise accuracy. Here, the class names were represented by the first three characters of a word. For class names with two words, the first character from the first word and the first two characters from the second word were used to represent the class name. Here, Unk denotes the unknown class.}
\label{fig:confMatESPNet}
\end{figure}

\begin{table}[t!]
\centering
\begin{subtable}[b]{\columnwidth}
\centering
\resizebox{0.9\columnwidth}{!}{
\begin{tabular}{l|c|ccccccccccccccccccc}
\toprule
\textbf{Network } & \textbf{mIOU} & \textbf{Roa} & \textbf{Sid} & \textbf{Bui} & \textbf{Wal} & \textbf{Fen} & \textbf{Pol} & \textbf{TLi} & \textbf{TSi} & \textbf{Veg} & \textbf{Ter} & \textbf{Sky} & \textbf{Per} & \textbf{Rid} & \textbf{Car} & \textbf{Tru} & \textbf{Bus} & \textbf{Tra} & \textbf{Mot} & \textbf{Bic} \\
\midrule
\textbf{ENet} \cite{paszke2016enet} & 58.29 & 96.33 & 74.24 & 85.05 & 32.16 & 33.23 & 43.45 & 34.10 & 44.02 & 88.61 & 61.39 & 90.64 & 65.51 & 38.43 & 90.60 & 36.90 & 50.51 & 48.08 & 38.80 & 55.41 \\
\textbf{ERFNet} \cite{romera2018erfnet} & \textbf{68.02} & 97.74 & 80.99 & 89.83 & 42.46 & 47.99 & 56.25 & 59.84 & 65.28 & 91.38 & 68.20 & 94.19 & 76.75 & 57.08 & 92.76 & 50.77 & 60.09 & 51.80 & 47.27 & 61.65 \\
\textbf{ESPNet (Ours)} & 60.34 & 95.68 & 73.29 & 86.60 & 32.79 & 36.43 & 47.06 & 46.92 & 55.41 & 89.83 & 65.96 & 92.47 & 68.48 & 45.84 & 89.90 & 40.00 & 47.73 & 40.70 & 36.40 & 54.89\\
\bottomrule
\end{tabular}
}
\caption{Class-wise comparison on the \textit{test} set}
\label{tab:compareClassSup}
\end{subtable}
\vfill
\begin{subtable}[b]{\columnwidth}
\centering
\resizebox{0.6\columnwidth}{!}{
\begin{tabular}{l|c|ccccccc}
\toprule
\textbf{Network} & \textbf{mIOU} & \textbf{Flat} & \textbf{Nature} & \textbf{Object} & \textbf{Sky} & \textbf{Construction} & \textbf{Human} & \textbf{Vehicle} \\
\midrule
\textbf{ENet} \cite{paszke2016enet} & 80.40 & 97.34 & 88.28 & 46.75 & 90.64 & 85.40 & 65.50 & 88.87 \\
\textbf{ERFNet} \cite{romera2018erfnet} & \textbf{86.46} & 98.18 & 91.12 & 62.42 & 94.19 & 90.06 & 77.43 & 91.87 \\
\textbf{ESPNet (Ours)} & 82.18 & 95.49 & 89.46 & 52.94 & 92.47 & 86.67 & 69.76 & 88.45 \\
\bottomrule
\end{tabular}
}
\caption{Category-wise comparison on the \textit{test} set}
\label{tab:compareCatSup}
\end{subtable}
\caption{Comparison on the Cityscape dataset. For comparison with other networks, please see the \textbf{Cityscape leader-board}: \url{https://www.cityscapes-dataset.com/benchmarks/}.}
\label{tab:compareClassCatSup}
\end{table}

\begin{table}[t!]
\centering
\resizebox{0.6\columnwidth}{!}{
\begin{tabular}{l|c|ccccccc}
\toprule
\textbf{Network} & \textbf{mIOU} & \textbf{Flat} & \textbf{Nature} & \textbf{Object} & \textbf{Sky} & \textbf{Construction} & \textbf{Human} & \textbf{Vehicle} \\
\midrule
\textbf{ENet} \cite{paszke2016enet} & 0.33 & 0.61 & 0.57 & 0.16 & 0.37 & 0.35 & 0.08 & 0.20 \\
\textbf{ERFNet} \cite{romera2018erfnet} & 0.25 & 0.73 & 0.29 & 0.16 & \textcolor{red}{0.03} & 0.23 & 0.06 & 0.24 \\
\textbf{ESPNet (Ours)} & \textbf{0.40} & 0.66 & 0.69 & 0.20 & 0.52 & 0.32 & 0.16 & 0.21 \\
\bottomrule
\end{tabular}
}
\caption{Category-wise comparison on the Mapillary \textit{validation} set. ESPNet learned generalizable representations of objects and outperformed both ENet and ERFNet in the wild.}
\label{tab:compareCatMapSup}
\end{table}

\end{document}